\documentclass{article}

\usepackage{PRIMEarxiv}

\usepackage[utf8]{inputenc} %
\usepackage[T1]{fontenc}    %
\usepackage{hyperref}       %
\usepackage{url}            %
\usepackage{booktabs}       %
\usepackage{amsfonts}       %
\usepackage{nicefrac}       %
\usepackage{microtype}      %
\usepackage{graphicx}
\usepackage{url}
\usepackage{xcolor}
\definecolor{newcolor}{rgb}{.8,.349,.1}
\usepackage{tabularx}
\usepackage{standalone}
\usepackage{textcomp, gensymb}
\usepackage{mathpazo}
\usepackage{longtable}
\usepackage[normalem]{ulem}
\usepackage{float}
\usepackage{svg}
\usepackage{refcount}
\usepackage[flushleft]{threeparttable}
\DeclareMathAlphabet{\pazocal}{OMS}{zplm}{m}{n}

\usepackage{amsmath,amsthm,dsfont,mathtools} %
\usepackage{pgf-pie}
\usepackage{tikz}
\usepackage{mwe}
\usepackage{subcaption}
\usepackage[hypcap=true]{caption}
\usepackage[figure,table]{hypcap}

\graphicspath{{media/}}     %

\def\BibTeX{{\rm B\kern-.05em{\sc i\kern-.025em b}\kern-.08emT\kern-.1667em\lower.7ex\hbox{E}\kern-.125emX}}
\definecolor{lightgray}{gray}{0.9}

\title{eXplainable Artificial Intelligence on Medical Images: A Survey}

\author{
	Matteus Vargas Simão da Silva, Rodrigo Reis Arrais, Jhessica Victória Santos da Silva \\
	Eldorado Research Institute \\
	Campinas, São Paulo - Brazil\\
	\texttt{\{matteus.silva, rodrigo.arrais, jhessica.victoria\}@eldorado.org.br}
	\And
	Felipe Souza Tânios, Mateus Antônio Chinelatto, Natalia Backhaus Pereira \\
	Eldorado Research Institute \\
	Campinas, São Paulo - Brazil\\
	\texttt{\{felipe.tanios, mateus.chinelatto, natalia.pereira\}@eldorado.org.br}
	\And
	Renata De Paris, Lucas Cesar Ferreira Domingos, Rodrigo Dória Villaça \\
	Eldorado Research Institute \\
	Campinas, São Paulo - Brazil\\
	\texttt{\{renata.paris, lucas.domingos, rodrigo.villaca\}@eldorado.org.br}
	\And
	Vitor Lopes Fabris, Nayara Rossi Brito da Silva, Ana Cláudia Akemi Matsuki de Faria \\
	Eldorado Research Institute \\
	Campinas, São Paulo - Brazil\\
	\texttt{\{vitor.fabris, nayara.silva, ana.faria\}@eldorado.org.br}
	\And
	José Victor Nogueira Alves da Silva, Fabiana Cristina Queiroz de Oliveira Marucci, Danilo Xavier Silva \\
	Eldorado Research Institute \\
	Campinas, São Paulo - Brazil\\
	\texttt{\{jose.silva, fabiana.marucci, danilo.silva\}@eldorado.org.br}
	\And
	Francisco Alves de Souza Neto, Vitor Yukio Kondo, Cláudio Filipi Gonçalves dos Santos \\
	Eldorado Research Institute \\
	Campinas, São Paulo - Brazil\\
	\texttt{\{francisco.neto, vitor.kondo, claudio.santos\}@eldorado.org.br}
}

\newcommand{\arrais}[1]{\leavevmode\color{black}{#1}}

\newcommand{\mateus}[1]{\leavevmode\color{black}{#1}}
\newcommand{\felipe}[1]{\leavevmode\color{black}{#1}}
\newcommand{\lucas}[1]{\leavevmode\color{black}{#1}}

\newcommand{\natalia}[1]{\leavevmode\color{black}{#1}}
\newcommand{\renata}[1]{\leavevmode\color{black}{#1}}

\newcommand{\vilaca}[1]{\leavevmode\color{black}{#1}}
\newcommand{\fabris}[1]{\leavevmode\color{black}{#1}}

\newcommand{\fabiana}[1]{\leavevmode\color{black}{#1}}
\newcommand{\akemi}[1]{\leavevmode\color{black}{#1}}
\newcommand{\jose}[1]{\leavevmode\color{black}{#1}}
\newcommand{\nayara}[1]{\leavevmode\color{black}{#1}}

\begin{document}
\maketitle

\begin{abstract}
Over the last few years, the number of works about deep learning applied to the medical field has increased enormously. The necessity of a rigorous assessment of these models is required to explain these results to all people involved in medical exams. A recent field in the machine learning area is explainable artificial intelligence, also known as XAI, which targets to explain the results of such ´black box´ models to permit the desired assessment. This survey analyses several recent studies in the XAI field applied to medical diagnosis research, allowing some explainability of the machine learning results in several different diseases, such as cancers and COVID-19.
\end{abstract}


\section{Introduction}
\label{s.introduction}

When it comes to artificial intelligence (AI) tasks, deep learning systems—exemplified by deep neural networks—are quickly becoming 
the industry standard ~\cite{das2020opportunities}. This includes everything from language comprehension and speech/image recognition to machine translation and 
planning, and even game playing and autonomous driving. Therefore, familiarity with deep learning is rapidly evolving from a specialized 
plus to a necessary requirement in many elite academic settings and a significant competitive advantage in the business world's job market. 
The "black box" concept, wherein Deep Neural Networks are said to lack transparency or interpretability of how input data are transformed 
into model outputs, is a major concern for the widespread application of Deep Neural Networks ~\cite{8466590, doi:10.1073/pnas.1900654116}. Many nonlinear, intertwined relations connect 
the various "layers" in a neural network. It is unrealistic to expect to understand the neural network's decision-making process even after 
inspecting all these layers and describing their relations. The lack of interpretability is causing growing concern across a variety of 
application domains because it can have far-reaching and unintended consequences. Medical imaging is one area where deploying AI models 
is met with skepticism due to the high stakes involved in a wrong classification ~\cite{8419428, VANDERVELDEN2022102470}. This paper reflects on recent investigations 
regarding the interpretability and explainability of Deep Learning methods.%

\subsection{Artificial Intelligence}
\label{ss.ai}

Artificial Intelligence (AI) is transforming the way real life 
experiences is addressed  by developing machines able to think like humans and mimic its 
behaviors including learning, reasoning, planning, predicting, and so on ~\cite{das2020opportunities}. 
Machine Learning (ML), which is a subset of Artificial Intelligence (AI), contains a set of algorithms 
with the ability of improving the performance of some task by experience to provide an inductive inference 
~\cite{mitchel1997machine}. These algorithms help machine to understand patterns within data and to develop expert systems 
in predicting or discoverying an unseen information. There are many ways that machines can understand these underlying 
patterns by using supervised, unsupervised techniques, neural networks and deep learning. In the supervised techniques 
the goal is to find a model from training data that can be used to predict/classify a target or a value of an unseen data 
based on input features. On the other hand, the goal of the unsupervised techniques is to find a pattern or discribing a 
set of data also based on a training data but without predicting an output attribute~\cite{TAN06}. %

\subsection{Explainable Artificial Intelligence (XAI)}
\label{ss.XAI}

While there is a general consensus that machine learning models should be easy to understand, there is a challenge on what exactly constitutes interpretability~\cite{Lipton2016mythos}. 
Models' interpretability has been defined in terms of their openness, accuracy, reliability, and ability to be understood ~\cite{Lipton2016mythos, Freitas2013comprehensible}. Many concepts of interpretability 
have been formulated within the context of computer systems, with little consideration given to the literature on interpretability that has been produced within 
the fields of the social sciences and psychology ~\cite{DBLP:journals/corr/Miller17a}. Accordingly, a common criticism of these definitions of interpretability is that they do not place sufficient 
emphasis on the user of interpretable machine learning systems. As a result, the produced models and explanations do not cater to the requirements of the target audience ~\cite{DBLP:journals/corr/Miller17a}.

\subsection{Explainable Artificial Intelligence in Medical Images (XAI)}
\label{ss.MIXAI}
\begin{figure}
\centering
\begin{minipage}{.5\textwidth}
    \centering
    \hspace{1cm}
    \includegraphics[scale = 0.15,bb= 0 0 754 276]{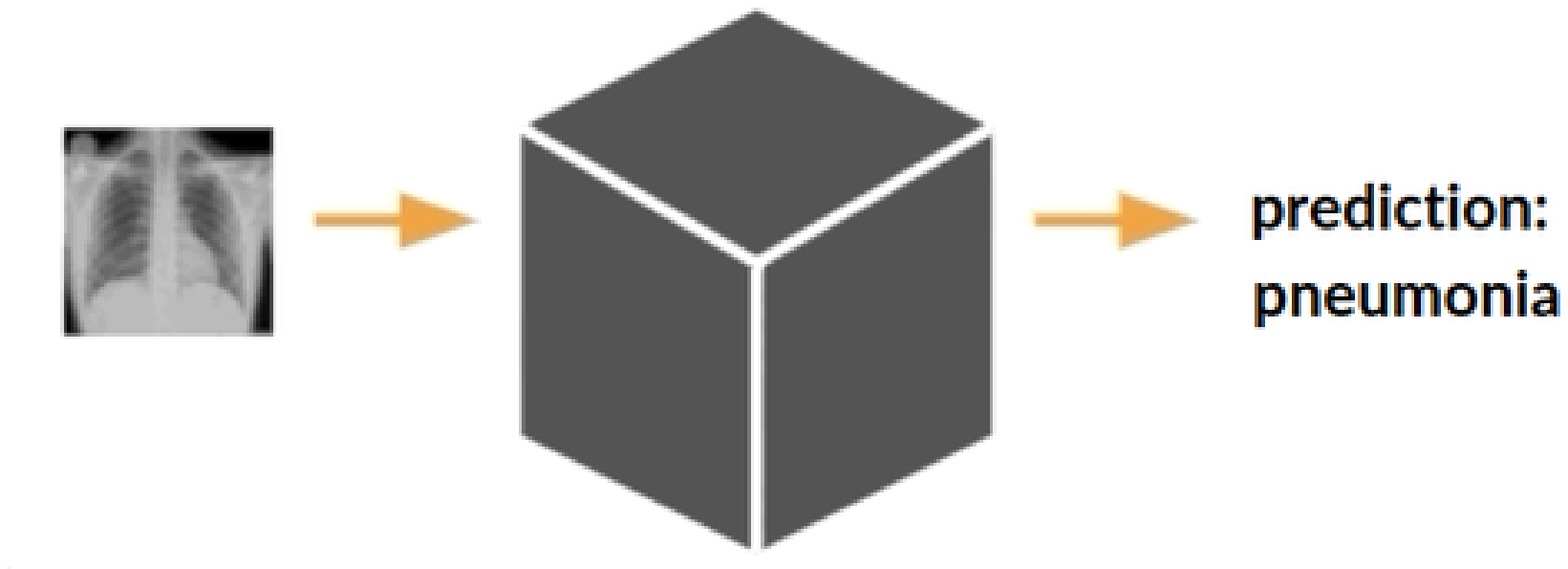}
    \caption{Black box concept illustration}
    \label{f.xai_1}
\end{minipage}%
\begin{minipage}{.5\textwidth}
    \centering
    \hspace{1cm}
    \includegraphics[scale = 0.15,bb=0 0 754 276]{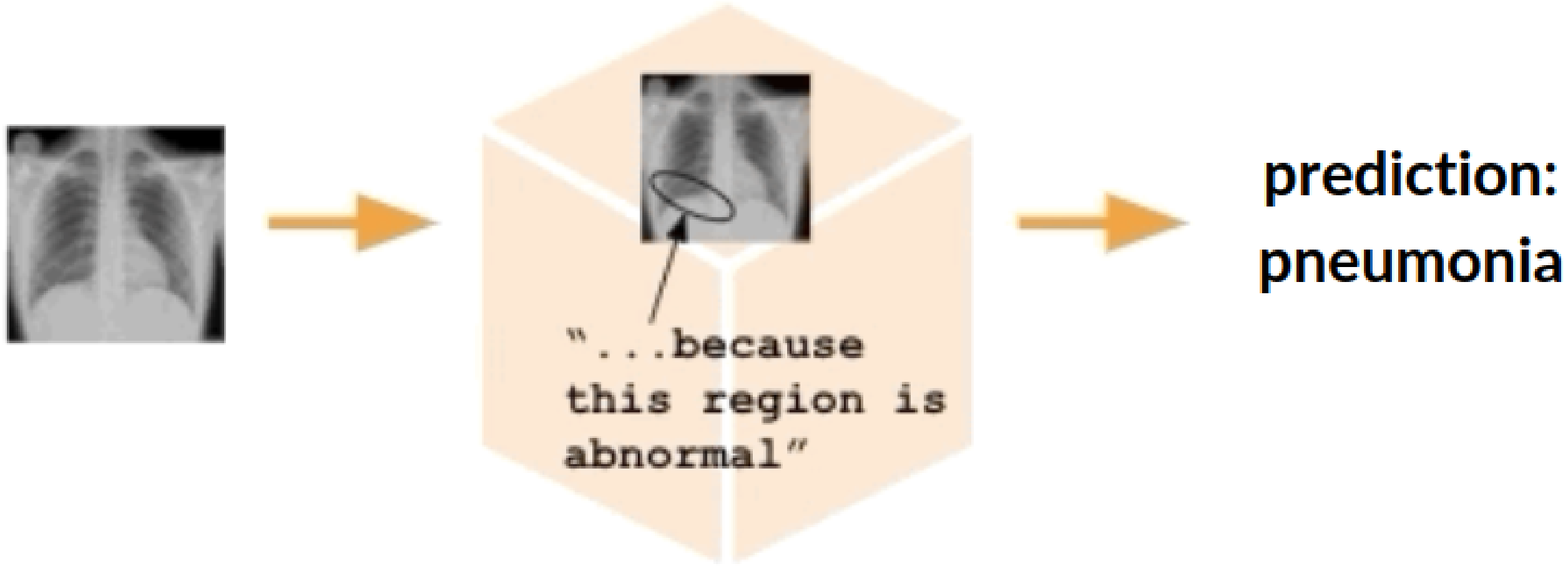}
    \caption{Explained black box illustration}
    \label{f.xai_2}
\end{minipage}
\end{figure}

Medical images are one of the most important clinical diangostic tool in 
medicine ~\cite{shung2012principles}. These images have properties that vary 
depending on the medical diagnosis and anatomical local such as skin~\cite{hu2022x,lucieri2022exaid,stieler2021skinimage,lucieri2020interpretability}, 
chest~\cite{lenis2020domain,hu2022x,brunese2020explainable,corizzo2021explainable,mondal2021xvitcos}, 
brain~\cite{bang2021spatio,li2021explainable} liver~\cite{MOHAGHEGHI2022105106}, and others.
Deep Learning algorithms have many critical applications in healthcare, including predicting patient risk of sepsis, search strategy and selection criteria, 
medical image, electronic health record, genomics, and others ~\cite{Yang2021review}. Thus, interpretable ML enables the end user to interrogate, comprehend, 
troubleshoot, and even enhance the machine learning system. In such cases, there is a high demand for interpretable ML models. End users, such as clinicians, 
may examine interpretable ML models before taking action ~\cite{8419428}. From the standpoint of physicians and patients, a model's output is not particularly relevant or 
accountable if it cannot be explained. An algorithm that detects pneumonia but cannot explain why a patient gets this diagnosis (Figure ~\ref{f.xai_1}) is less likely 
to be trusted and appreciated than a model that can provide some insight into its reasoning (figure ~\ref{f.xai_2}). Interpretable machine learning systems provide users 
with reasons to accept or reject forecasts and recommendations by explaining the logic behind them. 
Machine learning systems may be biased in their recommendations and decisions ~\cite{doi:10.1177/2053951715622512}. Interpretability is necessary to guarantee that these systems are bias-free and give 
individuals of all racial and socioeconomic backgrounds a fair rating ~\cite{10.1145/2939672.2945386}. Finally, tens of millions of people around the world are already benefiting from the decisions and 
recommendations made by machine learning algorithms, as seen on streaming services and social networks. These predictive algorithms are having far-reaching disastrous 
effects on society ~\cite{DBLP:journals/corr/Wang17j}, including the deskilling of professionals like doctors ~\cite{10.1145/2939672.2945386}. Given the difficulty of analyzing large amounts of healthcare data, the use of machine 
learning techniques to solve these problems is inevitable. However, there is an urgent need to establish uniform criteria for interpretable ML in this field.

\subsection{Scope of this work}
\label{scope-of-this-work}

The papers presented in this survey were selected based on the following criteria regarding their contributions and novelty:
\begin{itemize}
\item The application described in each paper should involve deep learning in some sense;
\item the data used had to be specifically medical images from varying fields of medicine;
\item  the publication year should be higher than or equal to 2020.
\end{itemize}

After filtering the retrieved works in regards to the aforementioned criteria, $32$ remained, notably, the ones reviewed in this survey. Table~\ref{tab:scope-of-work} describes them.

\begin{longtable}{|c|p{0.9\linewidth}|}
	\caption{Papers selected for this survey}
	\label{tab:scope-of-work} \\
	\hline
	Reference & Description \\ \hline
	\endfirsthead
	\endhead
	\cite{MOHAGHEGHI2022105106} & Improvement on CNN's performance for liver image segmentation through explainable techniques \\ \hline
	
	\cite{SHI2021100038} & Uses a region-attention module to locate important areas for the detections of a CNN on Geographic Atrophy in OCT volume scans \\ \hline
	
	\cite{QUELLEC2021102118} & Image segmentation technique coupled with a classifier in an end-to-end manner \\ \hline
	
	\cite{shi2021covid} & Saliency maps, severity assessment and prediction confidence for interpreting a teacher-student knowledge distillation network for COVID-19 diagnosis \\ \hline
	
	\cite{stieler2021skinimage} & Skin image classifier explanations through a novel perturbation method for the LIME framework \\ \hline
	
	\cite{hossain2020covid19} & 5G network integration for a ResNet50, Deep Tree and InceptionV3 for COVID-19 detection with GradCAM- and LIME-based visual explanations \\ \hline
	
	\cite{lucieri2022exaid} & Novel XAI framework based on Concept Activation Vectors and Concept Location Maps for mapping and highlighting human-understandable dermatology concepts \\ \hline
	
	\cite{calderon2021improving} & Improvement of uncertainty estimation for medical analysis reliability through the MixMatch algorithm \\ \hline
	
	\cite{ong2021comparative} & SqueezeNet were used to recognize lung diseases, and to give the explainability, LIME and SHAP were chosen. \\ \hline
	
	\cite{wang2021xai} & Guided Back-propagation for ultrasound images of knuckles and heart \\ \hline
	
	\cite{lucieri2020interpretability} & Concept Activation Vectors to interpret an Inception v4 learned concepts in comparison to the dermatology field \\ \hline
	
	\cite{corizzo2021explainable} & Grad-CAM visualization to assess potential lesions in Chest X-Ray image clusters based on a U-Net and a VGG16 feature extractions \\ \hline
	
	\cite{deramgozin2021hybrid} & Facial expression recognition CNN with LIME for region-of-interest visualization and an autoencoder-based Facial Actions Unit extractor \\ \hline
	
	\cite{raihan2022malaria} & Feature extraction framework involving a small CNN and Whale Optimization Algorithm and the use of SHAP for interpretability \\ \hline
	
	\cite{li2021explainable} & Explainable Ensemble Gaussian Kernel is created to compete against black box deep learning models using Random Forest Classifier. \\ \hline
	
	\cite{bang2021spatio} & Application of Layer-wise Relevance Propagation technique on a 3D-CNN trained on electroencephalogram data \\ \hline
	
	\cite{pianpanit2021parkinson} & Multiple explanation methods for interpretability of SPECT images classified by a PD Net and a Deep PD Net \\ \hline
	
	\cite{uehara2021explainable} & PixelCNN-based visualization network to interpret a CNN's diagnosis of pathological images from learned features \\ \hline
	
	\cite{mondal2021xvitcos} & Novel model based on the ViT-B/16 architecture to diagnose COVID-19 from CXR or CT images using Gradient Attention Rollout for explainability \\ \hline
	
	\cite{o2022explainable} & A possible XAI-involving cytoscopy extraction and classification is discussed \\ \hline
	
	\cite{painassessement2021} & Development of a mobile app for CNN-based assessments of neonatal pain, with Integrated Gradient algorithm for visualization of areas for the classifications \\ \hline
	
	\cite{hu2022x} & New similarity-based saliency maps for visual explanations of deep learning models and grouping of similar CXRs from a given query  \\ \hline
	
	\cite{li2020neural} & Novel explainable framework providing neural explanations and semantic image compression, tested on MNIST, CIFAR10 and Caltech256 datasets \\ \hline
	
	\cite{lenis2020domain} & Novel local XAI framework relating the CNN classification to the user's understanding of pathology \\ \hline
	
	\cite{rasaee2021modresnet} & ResNet- and Multi-Task Learning-based network with Grad-CAM in the last activation layers for explainability \\ \hline
	
	\cite{papadimitroulas2021artificial} & Review of recent developments in oncological radiomics, and a discussion of XAI-based techniques in medicine \\ \hline
	
	\cite{arefin2021non} & Classification of Macular Disease with a CNN trained on OCT images, with an expert's analysis of misclassified samples \\ \hline
	
	\cite{civit2020dual} & CNN coupled with a segmentation module to classify optic disc and cup images for glaucoma detection \\ \hline
	
	\cite{kabakcci2021automated} & Membrane Intensity Histogram calculation for immunohistochemistry breast tissue slides \\ \hline
	
	\cite{liz2021ensembles} & Heatmaps of a CNN ensemble highlighting the most influential and uncertain areas for pediatric pneumonia detection \\ \hline
	
	\cite{turkish9302311} & Visualization of the intermediate layers of a CNN for healthy neonatal thermograms classification\\ \hline
	
	\cite{brunese2020explainable} & Grad-CAM visualization of a VGG-16 activations for COVID-19 and pulmonary diseases identification \\ \hline
\end{longtable}

\subsection{Methodology}
\label{methodology}

This subsection describes the systematical analysis employed on the aforementioned papers in order to highlight possible improvements, state-of-the-art methodologies and manners in which to evolve the field of eXplainable AI. The fundamental points used to analyze the works presented in this survey are:

\begin{itemize}
	\item \textbf{Methodology proposed:} Analyzing the details of the methodology proposed in each paper guarantees its reproducibility, impact in the XAI research field and overall scope of each work's contributions, as well as open problems to be solved;
	\item \textbf{Datasets used:} The utilization of appropriate datasets for each task is essential to correctly evaluate obtained results and understand models' shortcomings. Therefore, analyzing the datasets used in each work is necessary to completely understand their contributions;
	\item \textbf{Evaluation method:} As important as the dataset applied; the evaluation method determines how the papers' results should be taken into consideration;
	\item \textbf{Results achieved:} With the previous points established, the results are then analyzed to understand each paper's impact on the field;
	\item \textbf{Explainability proposed:} The methods applied to interpret models' results and explain the impact of present features in the medical images.
\end{itemize}

\subsection{Survey Overview}
\label{paper-overview}

The reaminder of this survey is organized as follows: Section 2 describes the known XAI methods, 
its main development tools that assist to analyze and visualize the model interpretability, and provides a brief 
understanding on how explainability applied on Deep Learning algorithms for medical images is of paramount importance 
for diagnostics in medicine. Section 3 presents a brief description of the XAI methods used for 
analysing medical images based on Deep Learning for each of the 32 selected papers. Section 4 summarizes
the Deep Learning models, XAI methods, and the types of medical images found in the selected papers and provides a discussion
on the quantitative analysis. Section 5 concludes this study which led to this survey and gives 
direction for future works.
\section{Explainability}
\label{sec:explainability}

\lucas{As the machine learning application field grows in complex problem resolution, the need to understand and justify the acquired results is a latent problem. The algorithms produced can only be debugged or inspected when they are interpretable, which is a valuable feature during research and development stages, as well as after implementation~\cite{molnar2022}. The eXplainable Artificial Intelligence field comes as an approach to fill this gap. This field aims to produce visualizations, natural language, or mathematical equations that represent relevant knowledge from a machine learning algorithm, relating to any correlations found in the data or taught by the model~\cite{murdoch2019definitions}.} %

\lucas{According to~\cite{molnar2022}, the explainability can be achieved in different stages of the artificial intelligence pipeline: during the pre-prediction phase, answering the question of how the AI algorithm creates the model (Algorithm Transparency); during the prediction stage, understanding how the model makes predictions and which parts are affecting the results (Holistic Model Interpretability and Global Model Interpretability on a Modular Level); and post-prediction stages, giving insights on why the model predicted particular results for an instance or a group of instances (Local Interpretability for a Single Prediction or a Group of Predictions). With the increased notability that eXplainable AI received on the last years, various researches were conducted to develop more accurate methods in this area. The next section will cover some of the most common methods of explainable AI in the image area.}

\subsection{Known Explainability Methods}
\label{subsec:xai_methods}

\mateus{This section reviews known methods that are commonly used in the eXplainable AI field. The presented methods are extensively used out-of-the-box or as part of a more complex approach on the state-of-the-art techniques, discussed on Section~\ref{known_approaches}.}

\subsubsection{Grad-CAM method}

\mateus{Gradient-weighted Class Activation Mapping (Grad-CAM)~\cite{selvaraju2017grad} is a gradient-based method that aims to generate a localization map of the important regions in the image that contribute the most to the decision taken by the network. Since convolutional layers retain some spatial information, this method uses the gradient carried into the last convolutional layer to attribute an importance value for each neuron of the network on the decision taken.}

\mateus{One of its great advantages when comparing with other similar methods is that Grad-CAM requires no re-training or architectural changes and can be directly applied to several CNN-based models. It can also be combined with Guided Backpropagation via element-wise multiplication (Guided Grad-CAM) to generate high-resolution and class-discriminative visualizations.}

\subsubsection{Concept Activation Vectors}
\fabris{CAVs~\cite{kim2017interpretability} are an interpretability technique that generates global explanations for neural networks based on some user-defined concept~\cite{molnar2020interpretable}. A dataset that condenses the concept needs to be gathered, as well as a random one with lots of unrelated images. For a single instance, a binary classifier is trained on these two datasets with the task of classifying between the class of interest or not.}

\fabris{The CAV consists specifically of the coefficient vector of this binary classifier, and Testing with CAVs (TCAVs) allow us to average how many concept-based contributions, emanating from its own dataset, are related to the random images regarding the class of interest. In essence, with this technique, a high-level user-defined concept, supported by its dataset, is related to a class of interest positively or negatively.}

\fabris{With TCAVs, the user can define a high-level concept by just gathering data that relates to and represents it. This is very useful in the medical field, since medical specialists do not need to understand deeply the intricacies of neural networks: they can just collect data regarding the pathology or patient and apply TCAV, which will automatically relate (positively or not) the defined concepts with the existing classes.}%

\subsubsection{DeepLift}

\natalia{As described in~\cite{shrikumar2017learning}, Deep Learning Important FeaTures, also known as DeepLIFT, is an explainability method capable of defining contribution scores by comparing the difference of neuron activation to a reference behavior. According to the authors of~\cite{shrikumar2017learning}, DeepLIFT uses backpropagation to measure the contribution of each input feature when decomposing the output prediction. By analyzing the difference of an output when compared to a reference output, this method can compare the difference of an input to a reference input.}

\natalia{By using the difference-from-reference, DeepLIFT is capable of avoiding propagation issues when the gradient is zero or when the gradient has discontinuities. This approach is also capable of avoiding potentially misleading biases and of recognizing dependencies missed by other methods, since it is not affected by contribution saturation. When using DeepLIFT, it is important to keep in mind that choosing the reference input and output is crucial for achieving satisfactory results.}

\subsubsection{Saliency}

\fabris{Saliency Maps, introduced in~\cite{simonyan2013deep}, are a gradient-based visualization technique to understand the contribution of individual pixels of an image to its final classification by a neural network. This technique consists of applying a backward pass into the network and calculating the gradient of the loss function with respect to the input's pixels~\cite{molnar2020interpretable}. This way, it can retrieve the impact of each pixel in the backpropagation step and, consequently, how much it affects the final classification in regard to our class of interest.}

\fabris{It might interpret such results as another image, with the same size as our input image or at least easily projectable onto it, that indicates the most important pixels in our image to attribute it to class $c$.}

\subsubsection{Guided Backpropagation}

\jose{
Guided Backpropagation \cite{Guided_Backpropagation} is a combination of ReLU and deconvolution, in which the masked values of at least one of these is negative.

This method adds a guidance signal from the higher layer to usual backpropagation, preventing the backward flow of negative gradients, corresponding to the neurons which decrease the activation of the higher layer unit we aim to visualize.

The guided backpropagation works well without switches, allowing us to visualize the intermediate and last layers of a network.

}

\subsubsection{Layer-wise Relevance Propagation (LRP)}
\jose{Layer-wise Relevance Propagation (LRP) \cite{10.1371/journal.pone.0130140} is an explanation technique that gives explainability and scales to complex neural network models, with the capability of bringing results with different input modalities like text, images, and videos.  This technique works by propagating the prediction backward in the model, where the propagation garants what the neuron receives must be redistributed in equal amounts to the lower layers.

With the right set of parameters of the LRP rules, are possible complex models to obtain a high explanation quality.

}

\subsection{An Introduction to the Development Tools}
\label{subsec:xai_tools}

\jose{
Although the methods above achieve good results, their readability is not so simple. Since the output data is manually obtained, users may be in doubt about the confiability or compromise with the meaning of the data. To get around this, tools were created to analyze and visualize models.

The explainability models were widely requested, like in \cite{lucieri2022exaid}, and not exclusively by developers that needed to understand models, but also by users that wondered which evaluation criteria was being selected as a way of providing more reliable predictions. In addition to the transparency provided, in the case wrong evaluations would occur, with these frameworks it becomes easier to identify errors and successes and to enhance models' performance.

Furthermore, there are several frameworks that can be applied, but the ones selected in this survey were the most known by the community: SHAP and LIME. Moreover, Pytorch Captum was  also included  because it is interesting for their users since it is a native Pytorch framework. In the subsequent section, the aforementioned tools are explained.
}

\subsubsection{Pytorch Captum}

\jose{
Captum \cite{DBLP:journals/corr/abs-2009-07896} is a model interpretability and understanding library for PyTorch, containing general-purpose implementations of integrated gradients, saliency maps, smoothgrad, vargrad, and others.

Furthermore, Captum provides state-of-the-art algorithms, with an easy way to understand which features are contributing to a model’s output.
Captum has easy-to-use interpretability algorithms that can interact with PyTorch models. Also, it allows to quickly benchmark algorithms with others available in the library. The authors recommend the use of Captum mainly in models built with domain-specif libraries such as torchvision and torchtext because of their quick integration \cite{DBLP:journals/corr/abs-2009-07896}.

}

\subsubsection{SHAP}

\jose{

The Shapley value is a method used in game theory \cite{NIPS2017_8a20a862}, involving fairly distributing both gains and costs to actors working in a coalition. The Shapley value grants each actor a fair share depending on how much they contribute, meaning that its usage is correlated with a simple question: how much does something contribute to the results?

To solve this problem, the Shapley value derives from a formula:

$$w_1 \times MC_1 + w_2 \times MC_2 + \dots + w_i \times MC_i $$

Where $MC$ is the Marginal Contribution and $w$ is the weight of the marginal contribution. This formula is applied to every value in the model to find their individual contributions.

The SHAP (SHapley Additive exPlanations) \cite{NIPS2017_8a20a862} is a recent method based on the Shapley value that identifies and explains decisions made by AI models, decreasing black-box models' opaqueness. SHAP provides graphs and images of their explanations in an easy-to-understand manner.

}

\subsubsection{LIME}

\jose{
LIME (Local Interpretable Model-Agnostic Explanations) \cite{DBLP:journals/corr/RibeiroSG16} is a framework that explains predictions of any classifier in an interpretable and faithful manner. Its use provides graphs, tables, and explaining prediction of image objects with pros and cons.

When LIME receives a prediction model and a sample test, it provides faithful local explanations around the neighborhood of the instance being explained by applying perturbations to the image's superpixels and evaluating their impact on the final classification. By default it produces $5000$ samples, then it gets a target variable using the prediction model whose decisions it is trying to explain. With the substitute dataset created, LIME gets the weight of each line according to how close they are to the sample, ending with a selection of techniques (like Lasso) to obtain the most important samples.

LIME is limited to supervised models and is available in R and Python~\footnote{https://github.com/marcotcr/lime} with an API open code. It is also available other implementations~\footnote{https://paperswithcode.com/paper/why-should-i-trust-you-explaining-the}

}

\subsection{Explainability for Artificial Intelligent Solutions in Medical Images}
\label{subsec:xai_medical}

\lucas{It is well known that the correct analysis of medical images is a crucial part of the diagnosis. Also, the widespread availability of medical images requires more advanced tools to match the patterns found in multiple types of image sources (such as MRI, X-Ray, ultrasound, and others), analyze the nuances of each representation, and produce reliable conclusions about the data. These challenges required a significant improvement in the computational tools, as it required knowledge and application of the latest image manipulation methods~\cite{bankman2008handbook} to aid diagnostics in medicine.}

\lucas{Deep Learning became one of the most promising approaches for image analysis and processing, as the imaging challenges, such as the ImageNet Large Scale Visual Recognition Challenge (ILSVRC)~\cite{russakovsky2015imagenet}, have driven substantial advances in the area, pushing forward the field. Many applications benefited from the advances of the Deep Learning, drawing attention to machine learning applications too. Therefore, the employment of artificial intelligent solutions in the analysis of medical images is gaining notoriety and proving to be an adequate alternative to classic methods.}

\lucas{Interpretability, or explainability, is a necessary step in image classification for the medical area, as a decision in this field includes crucial risks and responsibilities. Giving such important decisions to machines that could not be held responsible would be comparable to absolutely avoiding human responsibility, representing an ethical issue and a fragility to malicious intents~\cite{tjoa2020survey}. \cite{holzinger2019causability} claims that not only explainability should be explored in the medical field, but also the causability. According to~\cite{holzinger2019causability}, causability is the degree to which a human expert can understand a statement's causal relationships effectively, efficiently, and satisfactorily in a given situation.} %

\section{Known Approaches}
\label{known_approaches}

\arrais{This sections reviews and briefly summarize some works that use methods based on deep learning applied to medical imaging and their respective methods of explainability.}

\subsection{Explainable AI and susceptibility to adversarial attacks: a case study in classification of breast ultrasound images}

\arrais{For the classification of breast ultrasound images, Rasaee et al. \cite{rasaee2021modresnet} investigated how some undetectable adversarial assault structures may help Convolutional Neural Networks explainability tools, such as GRAD-CAM \cite{Selvaraju2017gradcam}. For this work, the authors proposed a new network based on ResNet-50 and Multi-Task Learning (MTL) to improve breast ultrasound images' classification explainability and accuracy. The authors modified other aspects of the network, including a decoder box after the classifier, and six-sub boxes in the encoding box. To explain the model and visualize the feature maps, this study implemented GRAD-CAM as the XAI tool. To simplify, the explainability is applied only in the output layers - for the proposed ResNet-50, the majority of the XAI visualizations are related to the activation layer. Rasaee et al. used the “fast gradient sign method” \cite{Goodfellow2015attacks} to inject noise into the input image. The goal behind this idea is to consider scenarios where perturbations may distort the images such that the cancer location diagnoses changes, but the classification does not - this scenario could be very harmful, especially in situations like biopsy and resection. The authors handled two datasets. The first is a public dataset \cite{Dhabyani2020dataset} made of 780 breast cancer ultrasound images, recorded from 600 female patients (from 25 to 75 years old). These images are in the PNG format and have an average size of 500x500 pixels, categorized into benign, malign, and normal groups. The second dataset is made of 250 BMP breast cancer images, categorized into benign and malign images \cite{Rodrigues2017dataset}. This second dataset has two major problems for deep learning - the image size for feeding the model was varied and this dataset is not big enough for the model training. For these two aspects, the authors scaled all the images to 224x224 pixels and applied data augmentation.}

\arrais{The study brings a useful discussion about the presence of adversarial attacks in breast cancer images and its interference with cancer-type prediction, although the training and experiment stage could be more detailed, with more information about the hyperparameters and training adjustments.}
\subsection{Ensembles of Convolutional Neural Networks for pediatric pneumonia diagnosis (2021)}

\fabris{In this paper~\cite{liz2021ensembles}, the authors propose an ensemble of CNNs coupled with eXplainable AI techniques to simultaneously increase the robustness of CNN architectures with respect to low-data availability and quality, as well as provide interpretability for application in real medical domains. Heatmaps are created using the Keras Vis~\cite{kerasvis} package in order to highlight the most influential pixels of a classification instance. Six different CNN architectures are tested, with the highest-ranking model (consisting of 4 convolutional layers and 64 neurons in its FC layer) being trained on five different dataset folds and then used to compose the ensemble. Two datasets are used in this study: one with 950 low-res X-ray images of children's lungs (publicly unavailable) and one featured in Kermany et al.~\cite{kermany2018identifying} consisting of 5,856 high-res X-ray similar images. The results were compared to a state-of-the-art pathology detection X-ray model, CheXNet. 

The ensemble showed better results when compared to CheXNet's performance on the first dataset (0.89 average AUC to 0.76; 0.72 average TPR to 0.43) and stood competitive in the second (0.96 AUC to 0.94; 0.79 TPR to 0.886). The ensemble's heatmaps, achieved averaging each internal model's heatmaps, presented more significant information than their isolated counterparts. A measure of uncertainty is calculated with the standard deviation of each internal model's heatmaps, providing further aid to medical specialists when analysing the ensemble's results.}

\subsection{Dual Machine-Learning System to Aid Glaucoma Diagnosis Using Disc and Cup Feature Extraction}

\fabris{The work proposed in~\cite{civit2020dual} consists of a diagnostic aid tool for glaucoma detection in the form of a dual machine-learning model ensemble. The first part of this system applies segmentation techniques to independently detect and provide interpretability for both eye fundus' disc and cup, combining them and calculating each images' Cup to Disc Ratio (CDR) - a measure commonly used by specialists to diagnose glaucoma - and classifying examples on the basis of a CDR threshold. %

The segmentation subsystem has some extra components to it. Because these segmentation techniques can often lead to unreal disc and cup shapes, undermining the model's confiability, the resulting segmentation of images are passed to a RANSAC module to transform them into ellipsoids and calculate their original resemblance to such, a metric provided to the user as a segmentation adequacy evaluation. Moreover, the segmentation's resulting disc size is also evaluated and disproportional samples receive a less adequate final score. Both architectures are then combined. The segmentation part provides the final model with explicability (in the form of ellipsoids around discs and cups) and predictions; the classification part with only the latter. The authors do not specify the ensemble election method for most samples. In fact, they only stress that in medical fields, false negatives are the main concern and therefore if any of the above models classify a patient as having glaucoma, then the ensemble propagates such classification.

Two datasets are combined and used in this paper: RIM-One V3~\cite{rimone} (159 labeled and segmented eye fundus images) and DRISHTI~\cite{drishti} (101 similar images). They are statically and dinamically augmented for a resulting 9360-image dataset, split into 75\% for training and 25\% for testing. Histogram equalization and image resizing are also applied for use in both models. The U-Net's results are in line with other SOTA much bigger networks'. The comparison between their cups' Dice coefficient score (DCS) and SOTA's is 0.84/0.89 to 0.94/0.87 (RIM / DRI). For discs' DCS, 0.92/0.93 to 0.98/0.97. For MobileNet, the authors achieved 0.93 AUC, 0.89 ACC and 0.89 sensitivity (Se) - ResNet50, a much larger architecture, achieves 0.96 AUC, 0.90 ACC and 0.91 Se. However, combining the author's models, their results are 0.96 AUC, 0.88 ACC and 0.91 Se. This shows that the ensemble of those relatively small architectures compares easily with SOTA's results.}

\subsection{Automated scoring of CerbB2/HER2 receptors using histogram based analysis of immunohistochemistry breast cancer tissue images (2021)}

\fabris{In~\cite{kabakcci2021automated}, authors propose a method for the automated scoring of digitalized immunohistochemistry breast tissue slides for cancer detection in different scores (0 to 3). The images consist of tissue cells with immunohistochemistry staining in order to exalt tumor formation in cells. These images suffer a color deconvolution in two channels - hematoxylin and diaminobenzidine - and go through a cell nuclei and cell membrane detection, which ultimately results in the extraction of a 16-bin uniform Membrane Intensity Histogram (MIH) for every cell. The classification is performed on such histograms.

Three datasets are used in this paper. Two are novel publicly available datasets gathered from Istanbul Medipol University Hospital, namely ITU-MED-1 and ITU-MED-2~\cite{kabakcci2021automated}. The first is a balanced dataset of 13 cases with 191 tissue images and 62 thousand cells, and the second one, imbalanced as it is with 10 cases, 148 tissue images and 55 thousand cells, represents closely the real-world patient distribution. The third one is a publicly available contest dataset consisting of 79 IHC stained whole slide images~\cite{qaiser2018her2}.

Various classifiers are tested in this study. For the ITU-MED datasets, the highest ranking, both for cell-based and tissue-based scoring, is an Ensemble Boosted Tree, achieving 77.56\%/91.43\% (cell/tissue) accuracy on ITU-MED-1 and 91.62\%/90.19\% on ITU-MED-2. For the contest dataset from~\cite{qaiser2018her2}, a Cosine kNN achieved the highest results, ranking 6th among all 18 contest competitors. %
}

\subsection{Non-transfer Deep Learning of Optical Coherence Tomography for Post-hoc Explanation of Macular Disease Classiﬁcation}

\fabris{In~\cite{arefin2021non}, the authors discuss the necessity of big CNN models imbued with transfer learning techniques for classification of OCT images into four macular diseases. Such models are usually very large, with dozens of layers and tens of millions of parameters, which leads to a model opacity hampering the explicability of its filters and classifications. To this end, the authors propose a relatively small CNN (with only 5 convolutional layers) trained entirely on a public dataset of OCT images with 84,458 samples labeled between four disease classes (i.e. no pretraining). They analyze their model's 256 convolutional filters at the last layer, discovering that only eight of those didn't produce all-zero responses to input images - which indicates that most filters are redundant for OCT image classification.

The accuracy of the model proposed in this work is compared against other state-of-the-art (SOTA) transfer learning models in literature. The authors show that theirs achieve 97.9\% classification accuracy while the most accurate SOTA model achieves 97.7\%. Furthermore, the features learned in the last convolutional layer of their model show great capacity for disease differentiation without the need for the fully-connected layers at the end. The authors provide an explanation of the model's misclassifications, although on the form of a clinician's analysis of these OCT images rather than the application of automatic eXplainable AI methods.}

\subsection{Explainable Deep Learning for Pulmonary Disease and Coronavirus COVID-19 Detection from X-Rays}

\fabris{In~\cite{brunese2020explainable}, the authors propose the use of a deep learning model for Chest X-Ray (CXR) images classification regarding the Coronavirus disease. Since fast detection of true positive cases and accurate separation of false negatives and false positives are imperative to stabilize the epidemic, they consider automatic detection and classification of COVID-19 an important step. Therefore, a two-step methodology is proposed: firstly, a discrimination between healthy and pulmonary disease-ridden patients is done; lastly, if the patient is not healthy, a discrimination between general pneumonia and COVID-19 occurs. A VGG-16 architecture pre-trained on the ImageNet database is used in both cases, and fine-tuning is done. Moreover, to provide explainability to medical specialists who might be aided by such technology, a Gradient-weighted Class Activation Mapping algorithm (Grad-CAM), which indicates which areas of the input activated the network's neurons in order for it to arrive at its classification, is used.

Three datasets are combined in this work: publicly-available CXR images of COVID-19 cases; the dataset used in~\cite{ozturk2020automated}; and a dataset obtained from the National Institutes of Health Chest X-Ray, freely available for research purposes. %

The model that discriminates among healthy and generic pulmonary diseases achieves a sensitivity of 0.96 and specificity of 0.98, while its complementary model achieves, in those metrics, 0.87 and 0.94, respectively. Regarding accuracy, the first model achieves 0.96 and the second model, 0.98. The applied Grad-CAM correctly highlighted areas inside the chest cavity and, in particular, the lungs, overlapping with a radiologist's analysis on important areas for COVID-19 detection in CXRs.}

\subsection{Explainable AI and Mass Surveillance System-Based Healthcare Framework to Combat COVID-19 Like Pandemics}%

\felipe{The paper\cite{hossain2020covid19} proposes a health care framework to combat COVID-19-like pandemics. The framework is devided in three layers. The top Core Cloud layer stores and trains a deep neural network (DNN) model. The middle Edge Server layer uses 5G features, such as low latence, high data rates and the support of IoT devices to send the data to the Core Cloud layer and download the trained model, so it can run the new data through the already trained DNN model for Covid19 classification and explain the results. The bottom layer, which consists of the hospital and homes, receives both the results of the Edge Server classification and its explanation. The explanation method proposed is a knowledge map that uses Local Interpretable Model-Agnostic Explanations (LIME)~\cite{lime} and a visualization algorithm that uses gradient-weighted class activation mapping (Grad-CAM), generating a visual and attention-based explanation.}

\felipe{The paper compares only the results of DNN models run in a cloud layer and the Edge server. To test the model, the authors have used a representation of a edge server along with a mobile device, connected in 5G. To assure this connection, there was a python server checking the protocol messages exchanged. The edge server was an NVIDIA Jetson T2. The mobile used was a Samsung Galaxy S20 Plus, 128GB smartphone running Android version 10.  They have used three DNN models: ResNet50~\cite{resnet50}, deep tree, and Inception v3~\cite{inceptionv3}. The dataset selected consists of 2000 healthy samples, 2000 pneumonia samples (non-COVID-19), and 200 COVID-19 samples in the training set; the testing set consisted of 200, 200, and 50 samples, respectively. The origin of the samples is unclear, altough the paper suggests the existence of Kaggle and Github repositories for hosting chest X-ray and CT scan images. The training data was augmented using rotation, horizontal and vertical scalling. The augmentation rate is also unclear, although they report the augmentation parameters boundaries. The paper reports the best results when using the Mobile and cloud, instead of mobile and edge or only mobile. The parameters used were the latency and images in queue for processing. There is no mention to the explanation model tests nor its results.}

\subsection{Towards Domain-Specific Explainable AI: Model Interpretation of a Skin Image Classifier using a Human Approach}%

\felipe{This work~\cite{stieler2021skinimage} proposes a explanation model based on Local Interpretable Model-agnostic Explanations (LIME) that uses the dermatology ABCD-rule approach. To do this, instead of occluding areas (LIME's usual approach), it proposes a model that adds perturbations to the images by shifting and rotating, which are medically irrelevant, and altering the boundary and color (B and C of ABCD rule). The boundary perturbation is either negative, extracting the border area of the segmentation and drawing a line around the lesion, or positive, adding a blur on the lesion's edge. The color perturbation can also be negative by uniforming the color or positive by adding random color patches. The study simplifies the problem space by reducing the possible classes to only two: nevus (1354 samples) and melanoma (216 samples).}

\felipe{The paper proposes three different hypothesis:
\begin{itemize} 
    \item A: that the prediction for nevus will decrease with positive perturbation,
    \item B: that the prediction for nevus will increase with negative perturbation,
    \item C: that the model depends on the medically irrelevant perturbations. 
\end{itemize}To test these hypothesis, it uses the HAM10000 dataset~\cite{ham10000}, used in the ISIC Skin Lesion Classification Challenge, and a pre-trained MobileNet model with skin image data. Along with the F1-scores of the classifier, which are not relevant to the explainability of the model, it presents some empirical results using the true and false positive classifications. With the true positive results, it argues towards accepting hypothesis A and C, but not B. On the other hand, with the false negative results, it accepts only A and rejects B and C. It concludes showing how those results, even though trying to follow the ABCD rule, still need a translation from result to explanation for a final user. It also shows possible future works on the perturbations used and a feature dimension used to train that is relevant to the ABCD rule.}

\subsection{Developing an explainable deep learning boundary correction method by incorporating cascaded x-Dim models to improve segmentation defects in liver CT images }%

\felipe{This paper~\cite{MOHAGHEGHI2022105106} describes an explainable method to improve the CNN model for liver image segmentation. The approach was to first identify how human specialists perform this task and create a mechanism that uses human's method for validate a liver segment boundary, in order to correct the boundaries of a segment that was the output of a CNN. }

\felipe{The method implemented extracts the boundary of the initial 3D segmented image and, for each slice that has already a boundary, determine which are the boundary points and get feature array of those points. With those informations, it determines if the boundary points are valid or not using a 1D 4 layer dense model, which they call Boundary Validation, and correct the areas using a modified 2D U-Net, a 2D deep CNN model, which they call Patch Segmentation. Finally, they replace the invalid borders with the corrected generated images. To test the model, two public datasets, Silver07~\cite{silver07} and 3D-IRCADb~\cite{3d-ircad} with 20 volumes were used. Other 93 volumes were used as well, but they were private. This method of correcting after being segmented improved almost every model. However, the performance is dependent of the initial slice segmentation.}

\subsection{Explainable Features in Classification of Neonatal Thermograms}

\felipe{This paper~\cite{turkish9302311} investigates the high classification performance of neonatal thermograms. The analysis of neonatal thermograms is a technique described in the 1980s, however it has been explored further due to the deep neural network models for image analysis. This paper describes a model with a good classification performance (94.73\% accuracy) of a 6-layer convolutional deep neural network, with dropout layers~\cite{dropoutref} in between as well as two fully connected layers in the end for classification. The inputs are neonatal thermograms, which are colorful images with a resolution of 640x480 pixels. A novel and private dataset of 190 thermal images from 19 healthy neonates and 19 neonate patients was used. It also describes a data augmentation of this dataset by varying luminance, contrast, resolution, color, and by adding noise and shifting or flipping the image. It is not clear what the augmentation rate of the study was.}

\felipe{On the analysis, the paper focus on the visualization of the intermediate layers of the convolutional network to determine what was used in the classification. It points out that even though the layers were used the thermograms image correctly, focusing mainly on the neonate, the network used learned that patients usually had incubators, respiratory support devices, and measurement equipments, among other undesirable materials that had a big influence on the learning procedure. Therefore, the network did not rely mainly on the neonate itself.}

\subsection{Interpreting Skin Lesion Classifiers using CAV's}%

\lucas{The paper~\cite{lucieri2020interpretability} uses the problem of classification of skin diseases to understand what the DNNs learn and how the predictions are formulated in their internal structures. To understand the reasoning behind the predictions, the authors use Concept Activation Vectors (CAV), applied to identically distributed data, mapping the concepts learned by the DNN in latent space to human-understandable concepts in the dermatology area. First, the activations are extracted from the target layer of the chosen DNN and used to compute the CAV, and then the TCAV (Testing with CAV's) score is used to evaluate the importance of the given concept concerning the target class.}

\lucas{The results reported were obtained using one of the RECODs Lab (REasoning for COmplex Data~\cite{menegola2017recod}) base models, with Inception v4~\cite{szegedy2017inception} architecture, trained using the PH²~\cite{mendoncca2013ph} dataset and Seven-Point Checklist Dermatology dataset (derm7pt)~\cite{kawahara2018seven}. The results show strong correlations between the concepts usually applied in the dermatology field and the ones learned by the DNNs representations. Although the accuracies achieved were not high, mostly due to the CAV's computation process requiring the use of linear classifiers to obtain the normal vector on the decision hyperplanes, the TCAV score values, which quantify the positive or negative influence of the concepts on a specific target class, showed that exists an agreement between the reasoning of the DNN's models and the human specialists on the problem of classification of skin lesions.}

\subsection{USRNET}%

\lucas{In the paper~\cite{wang2021xai}, the authors proposed a method to achieve feature detection for ultrasound images based on the Deep Unfolding Super-resolution Network (USRNET~\cite{zhang2020deep}). To introduce the explainability, they use an XAI approach focused on guided back-propagation to detect and extract features from super-resolution neural networks, in particular the USRNET. When the feature gradient maps are obtained, a thresholding method is applied to achieve binarization of the points in the original map.}

\lucas{The data used in this study was obtained from ultrasound images of knuckles and heart, collected using the Clarius handheld ultrasound device. The experimental data was obtained from adjacent frames of an ultrasound video, using the same device. The results were compared with four state-of-the-art models, named SIFT~\cite{sift}, SURF~\cite{surf}, FAST~\cite{fast}, and ORB~\cite{orb}. The USRNET approach with guided back-propagation showed to be effective when detecting feature points located at high-frequency regions and, when the matching of the feature maps was computed using BRIEF descriptor and a brute-force matcher, it outperformed ORB and SIFT in terms of matching accuracy. The main contribution was the detection and explainability of high-frequency areas of the image, therefore, the next step is to extend this method to the entire image.}

\subsection{Decision support in medical healthcare}%

\lucas{As the COVID-19 pandemic increased the need for tools to assist in medical image analysis, an explainable system able to help physicians and clinicians to make a diagnosis is proposed by~\cite{corizzo2021explainable}. The system was not designed to make decisions, but aimed to cluster images from Chest X-Ray (CXR) from patients with pneumonia caused by COVID-19, highlighting the regions of interest from them. The system is composed of two main phases, the first is an adversarial learning component, with a U-Net generator and a VGG16 discriminator, and the second is composed of clustering, dimensionality reduction, and visualization techniques. Combining clustering and activation maps, the system can provide a visual explanation of the results.}

\lucas{The first component of the system is essentially used to provide the inference explanation, as a result of the adversarial output fed into the Grad-CAM method~\cite{selvaraju2017grad}, and the features used on the visualization, provided by the U-Net generator feature extraction layers. Those features are fed into the second stage, the clustering and visualization phase. This step aims to identify concepts in the data features, through three different clustering algorithms: k-Means, Birch, and Agglomerative. They are used to provide a diverse perspective of the features and also provide a redundancy to increase reliability. The resulting visualization is generated by the t-SNE algorithm~\cite{van2008visualizing}.}

\lucas{The dataset used in this study contained 3876 images of Pneumonia lesions and 150 images of COVID-19. The source of the data was not provided. Although the authors reported high values for precision~(0.840), recall~(0.829), and F-Score~(0.828) in the discriminator stage of the adversarial phase, the prediction confidence obtained varied in the predictions, pointing to the variability of the COVID-19 lesions. The clustering obtained by the different methods converged, showing an agreement between the methods, and the Grad-CAM representations highlighted the localization of the lesion and its degree of severity. In short, this paper presents a complete pipeline to aid clinicians in the diagnosis of COVID-19, showing the potential areas of the CXR where the lesion is located. Furthermore, it successfully used t-SNE and Grad-CAM to help with the visualization of the prediction provided by the method.}

\subsection{A Hybrid Explainable AI Framework}%

\lucas{A framework composed of a Convolutional Neural Network along with the application of the LIME technique was used in~\cite{deramgozin2021hybrid} to address the problem of Facial Expression Recognition (FER). The solution used a black box approach based on a 6-layer CNN to classify the input images according to their facial expressions and used the LIME technique to allow the visualization of active regions in the CNN. Also, a Facial Actions Unit (FAU) extractor, based on an Auto Encoder, was used on the input image to extract the action units and provide an extra factor to interpret both LIME and CNN results. In addition, the AUs generated by the FAU were used on an MLP classifier, resulting in a redundant classifier to reinforce or undermine the results of the first pipeline.}

\lucas{Using the CK+ dataset~\cite{ckplus} as input and the Openface~\cite{openface} tool as ground truth for the AUs used in the auto-encoder, the resulting pipeline obtained results that showed a competitive accuracy over the other state-of-the-art models. The results from the CNN surpassed the eXnet~\cite{riaz2020exnet}, the SOTA method with higher accuracy reported at this work by 0.75 percentual points, and the results from the redundant pipeline (FAU + MLP) were 8.75 percentual points worst than eXnet. On the explainability of the results, the intensity diagram for each AU extracted in this work was consistent with the emotions reported in the \emph{Facial action coding system}~\cite{ekman1978facial}, one of the central references of this area. One of the major contributions of this work is that the redundant part can be beneficial to aid the recognition, although not necessary as the explainability and classification results are better obtained from the functional pipeline. Also, as the next step to improve the results, the AUs list expansion can be performed to obtain better representations of the explainable part.}

\subsection{Explainable AI for COVID-19 diagnosis on CXR Image}%

\lucas{The work proposed by \cite{ong2021comparative} applied the SqueezeNet~\cite{ucar2020covidiagnosis} model to recognize pneumonia, COVID-19, and typical lung images on X-ray scans. They also suggested an explicable pipeline using the LIME and the SHAP approaches to enhance the interpretation of the pipeline results.}

\lucas{COVIDx dataset~\cite{wang2020covid} was used in this research due to its availability and representativeness. It consists of distinct, publicly modified datasets with 19,843 X-Ray images in total (8851 typical cases, 6069 pneumonia cases, and 4923 COVID-19 cases). Also, a data augmentation pipeline was introduced, aiming to reduce overfitting and improve the regularization of the model through rescaling, shearing, zooming, and flipping horizontally procedures applied to the input images. The experiment results achieved an overall accuracy of 84.3\%, however, the highest accuracy was obtained for the typical lung prediction due to its highest number of related images in the training dataset. Regarding the explicability, both SHAP and LIME were able to indicate the image regions used on the predictions, nevertheless, the visual inspection of the SHAP results shows that this approach went better in identifying the most relevant lung regions. This paper claims to be the first known application of SHAP for explicability on COVID-19-related issues, however, the authors suggested that they could achieve an accuracy improvement by employing a fine-tune on the SqueezeNet model.}

\subsection{Spatio-Spectral Feature Representation for Motor Imagery Classification Using Convolutional Neural Networks}%

\mateus{The paper~\cite{bang2021spatio} aims to obtain a novel method for the feature representation of electroencephalogram (EEG) data on motor imagery-based brain-computer interfaces (BCIs). The authors propose a method to create a filter-set that can generate a spatio-spectral feature representation that preserves the multivariate information of EEG data. That feature representation can then be presented to a 3D convolutional neural network that performs the classification to identify the user intent.}

\mateus{To validate the spatio-spectral characteristic and visualize the representation and results obtained by the filter and the 3D-CNN, the authors adopted the Layer-wise Relevance Propagation (LRP) XAI method. The LRP was adapted into a three-dimensional domain and used to create heatmaps of the relevance of EEG signals and then plot the topography of the brain by mapping the location of each EEG channel.}

\mateus{Experiments with LRP showed that the proposed method was able to maintain the spatio-spectral characteristics of the EEG data and give an intuitive understanding for the 3D-CNN decisions. The model also outperformed the competing state-of-the-art techniques obtaining a mean accuracy of 87.15\% (±7.31), 75.85\% (±12.80), and 70.37\% (±17.09) for the BCI competition data sets IV\_2a, IV\_2b \cite{tangermann2012review}, and OpenBMI data ~\cite{lee2019eeg}, respectively.}.

\mateus{Although this work does not present a novel approach or focuses specifically on explainable AI methods, it uses stablished techniques to visualize the process, improve interpretability and validate the obtained result.}

\subsection{Parkinson's Disease Recognition Using SPECT Image and Interpretable AI: A Tutorial}%

\mateus{The tutorial~\cite{pianpanit2021parkinson} presents a procedure to choose a suitable interpretation method for deep learning Parkinson's Disease (PD) recognition models based on single-photon emission computed tomography (SPECT) images. It compares a traditional classification method based on experts' analysis and a support vector machine (SVM) classifier with two simple neural networks (PD Net and Deep PD Net), alongside the following explainable AI's methods: Direct backpropagation (Saliency map), Guided backpropagation, Grad-CAM, Guided Grad-CAM, SHAP and DeepLIFT.}

\mateus{The public SPECT image dataset from the Parkinson's Progression Markers Initiative (PPMI)~\cite{marek2011parkinson} database was used. A total of 607 subjects, 448 with PD and 159 without, were used, with splits for training, validation and testing following the 80:10:10 ratio.}

\mateus{The results suggest that the combination of neural networks and interpretation models can outperform the methods based on experts' analysis and SVM. The possibility of overfitting was the most notable drawback. Guided back propagation showed the highest Dice coefficient and lowest mean square error and SHAP provided the best quality heatmap, better discriminating the difference between PD and NC subjects.}

\mateus{The main contribution of this tutorial is the demonstrated process to select a suitable interpretation method for a specific task with the help of interpretable and explainable AI methods, that can be applied to other tasks and problems.}

\subsection{Explainable Feature Embedding using Convolutional Neural Networks for Pathological Image Analysis}%

\mateus{The paper~\cite{uehara2021explainable} proposes an interpretable diagnosis method for the analysis of pathological images based on two combined neural networks, one focused on the diagnosis itself and the other on the visualization of the learned features from the inputs.}

\mateus{The Diagnosis network is a convolutional network that embeds the input images to a feature space, maps those extracted features to sequences of discrete variables corresponding to pathological features, and then classify the image based on the presence of the mapped pathologiacal features. The Visualization network is a PixelCNN~\cite{van2016conditional} model that can generate clear images that represent each of learned pathological features.}

\mateus{The method was first validated on the Kylberg texture dataset~\cite{kylberg2011kylberg} since textures images share similarities with pathological images, with the accuracy achieved being 99.3\%. The authors then applied the model to real pathological images of a uterine cervix provided by Nagasaki University Hospital, Japan, that were annotated by specialists. The dataset used contained 114,081 normal and 14,374 abnormal image patches.}

\mateus{The proposed approach was compared with authors' previous works~\cite{uehara2019prototype}, a prototype based network~\cite{li2018deep} and an Inception v3 model for cancer detection~\cite{liu2017detecting}. Comparing the area under the ROC curve (AUC), only the Inception v3 network yielded a better result (AUC=0.934) than the proposed method (AUC=0.928). However, the inception model does not provide explainable and interpretable results.}

\mateus{The proposed two network architechture containing a dignosis and a visualization network presents great advantages when compared to commom black box models. The explainability that the visualization network brings improves the reliability of the dignoses performed by convolutional networks and is one of the main contributions of this paper.}

\subsection{An explainable ensemble feedforward method with Gaussian convolutional filter}%

\mateus{With the objective of creating a transparent explainable method to rival the blackbox deep learning models, the authors propose a model, XEGK (Explainable Ensemble Gaussian Kernel)~\cite{li2021explainable}, based on the development of an explainable Gaussian Kernel (XGK). The XGK uses Gaussian Mixture Models to extract features from patches of the input images, producing both deterministic and stochastic mappings for feature representation. The XEGK combines those mappings to get a feature representation for the input. The paper contains the whole mathematical derivation of the method and presents the pseudocode for its implementation, both for single and multi-channel cases.}

\mateus{The authors investigated the performance of the XEGK on two datasets. Mitosis: composed of the challenging ICPR 2012 Mitosis dataset~\cite{ludovic2013mitosis} and TPAC 2016 dataset~\cite{veta2016tumor}, with two categories of cell, the mitosis and normal cell, each of them with 1253 images. And Brain Tumor: 3064 T1-weighted contrast-enhanced images with three kinds of brain tumor: glioma (1426 slices), meningioma (708 slices) and pituitary (930 slices).}

\mateus{Classification experiments were performed for single and multi-channel images, and results were compared with RAM~\cite{mnih2014recurrent}, VGG19~\cite{simonyan2014very}, LeNET-5~\cite{lecun1998gradient} and EPF-RF~\cite{pintelas2020explainable}, using Accuracy, Recall, Precision, and F-score. The setup with XEGK plus a Random Forest for classification outperformed all other methods on both datasets, achieving accuracies of 92.7\% and 91.3\% on Brain Tumor and Mitosis datasets respectively. It is interesting to notice that a statistical model combined with a classical machine learning algorithm is capable of rivaling stablished deep learning models with thousands or millions of parameters, but in a completely transparent way. The results on Mitosis dataset were also analyzed with SHAP.}

\subsection{Malaria cell image classification by explainable artificial intelligence}%

\mateus{The paper ~\cite{raihan2022malaria} proposes an explainable framework to detect malaria infected cells from grayscale cells images. The framework is composed by a combination of machine learning algorithms to extract and select relevant features from the images, followed by an XGBoost classifier~\cite{chen2016xgboost} to identify the parasitized cells. In order to interpret the model results and the contribution of each selected feature, the authors used SHAP.}

\mateus{The malaria dataset~\cite{rajaraman2018pre} is balanced, containing 27,558 cell images, 13,779 parasitized and 13,779 uninfected. In their research, the authors resized the images to 120x120 pixels and converted them to grayscale.}

\mateus{The main contribution of this paper is its feature extraction framework and the later analysis using SHAP: it first applies a Wavelet Packet Decomposition (WPD)~\cite{huang2008wavelet} with biorthogonal 6.8 wavelets to reduce the image size and extract features, then it feeds the "approximate" subband to a small CNN. This CNN was trained for classification, however only the convolutional layers and first fully connected layer are used to encode the features. The encoded output is analyzed using the Whale Optimization Algorithm (WOA)~\cite{mirjalili2016whale} and the optimal set of features is extracted. The XGBoost model achieved an accuracy, precision, recall, and F1 score of 94.78\% , 94.39\% , 95.21\%, and 94.80\% using the features extracted from with this framework. Other studies in this field have achieved higher accuracies, up to 99\% ~\cite{rajaraman2018pre, 9093853, rajaraman2019performance, 8697909}, however they are more complex and not explainable.}

\subsection{Explainable artificial intelligence (XAI): closing the gap between image analysis and navigation in complex invasive diagnostic procedures}

\natalia{Cystoscopy is widely used for bladder cancer detection and, as it is considered less invasive than surgical procedures, the article~\cite{o2022explainable} discusses the possibility of cystoscopy being a safest starting point for a semi-automated medical procedure. Since the 10th most common form of cancer worldwide is bladder cancer~\cite{bray2018global} cystoscopy plays an important role regarding periodic monitoring of the diseases. However, this type of cancer is described as heterogeneous and different in pathology and in molecular background, which can cause difficulties to detect lesions with standard white light cystoscopy, especially because this procedure requires a professional to conduct and interpret the images. With this background, the authors are convinced that an improvement in the navigation will enable automation, which will ensure reproducibility.}

\natalia{A major downside of cystoscopy is how painful it is, even flexible cystoscopy is considered uncomfortable for patients. In addition, the findings from this procedure can be often difficult to interpret and classify, which depends directly on the experience of the examiner. To balance this situation, the authors of the article~\cite{o2022explainable} suggest a combination of robotic cystoscopy with XAI-enhanced imaging, since it would require less experience to be done, increase patient comfort, shorten the procedure duration and improve diagnostic accuracy. The authors emphasize the importance of explaining to the patient that the procedure is not performed only with an autonomous robot, but with a human guidance. In this scenario, a trained nurse would insert a flexible XAI robot-assisted cystoscope and the semi-autonomous procedure would be performed quicker. It is important to point out that if the patient feels any kind of discomfort, they can interrupt the procedure at any time by telling the nurse or pushing an emergency button (patient-in-the-loop). Therefore, if this new procedure is performed obeying regulations rules, using a semi-automated XAI assisted cystoscopy has the potential to lead to more efficient diagnostics with interpretable results, and to increase patient safety.}

\subsection{xViTCOS: explainable vision transformer based COVID-19 screening using radiography}%

\natalia{The contamination speed of COVID-19 has created the need for faster forms of diagnosis. The article \cite{mondal2021xvitcos} mentions that the RT-PCR test is reliable, but has a long waiting time for the result. Based on this need, a new form of diagnosis was suggested with the use of Deep Learning in X-ray and computed tomography (CT) images of the lung. This idea surged because, according to the authors of \cite{mondal2021xvitcos}, the lungs of patients infected with COVID-19 have a ground glass appearance, which can be visually detected in X-ray and CT exams. However, most studies involving this type of diagnosis use convolutional neural networks (CNNs), which can be computationally expensive and do not usually have an overview of the image, which can generate interpretation bias.}

\natalia{In an attempt to overcome this problem, the article \cite{mondal2021xvitcos} aims to use Vision Transformer (ViT) for COVID-19 classification with a multi-stage transfer learning. The model, called xViTCOS, is referred as xViTCOS-CT when used on CT images, and as xViTCOS-CXR when used on X-ray images. The study chose the ViT-B/16 architecture with 16 x 16 pacth sizes. A multi-stage transfer learning approach was used to improve the model’s performance. It was first trained on ImageNet-21k dataset and finetuned on ImageNet-2012. Finally, the model was also finetuned on a chest radiography database.}

\natalia{To evaluate the study’s performance, a benchmark approach was developed by the authors of \cite{mondal2021xvitcos} to compare Accuracy, Precision, Recall, F1 score, Specificity and Negative Prediction Value. xViTCOS-CT had the best accuracy and F1 score among all, it also had high values of recall, precision and F1 score. xViTCOS-CXR performed even better, with the best accuracy in comparison to others and high values of recall, precision and F1 score. The explainability is shown with the Gradient Attention Rollout algorithm. The results of the algorithm were compared with radiologist’s interpretations and both models were able to correctly classify all regions and diagnosis.}

\subsection{NICE}%

\renata{The paper Neural Image Compression and Explanation (NICE) ~\cite{li2020neural} shows a deep learning based framework that integrates neural explanation and semantic image compression into a end-to-end training pipeline. To evaluate the performance of NICE, the authors compared the results by training the NICE pipeline and explaining the target using three image classification benchmarks: (1) MNIST dataset pretrained with LeNet5 classifier, CIFAR10 dataset pretarined with VGG11, and Caltech256 dataset pretrained with ResNet18 classifier.}

\renata {Experiments show that NICE is about 23x faster than Saliency Map, 16.5x than CAM and 2.8x faster than RTIS. Further, when we compared with other XAI methods, the sparse masks generated by NICE show more concise, coherent and match well with how humans explain their own predictions, specially in images with low resolution as MNIST and CIFAR10 images where NICE shows more concise and the boundaries of salient regions are much sharper than Saliency Map and RTIS.  }

\subsection{Classifier Interpretation by conterfactual impact analysis framework}%

\renata{In this paper~\cite{lenis2020domain}, the authors present a new framework for real time pathology classifier interpretation based on local explanations that provide potential insights for clinical application by relating the artificial neural networks algorithmic outcome to the user's understanding of pathology and reinforcing confidence in the prediction. Experiments were performed using the MobileNet classifier for distinguishing between healthy and scans with masses in the Database for Screening Mammography (DDSM)~\cite{heath00} and the Curated Breast Imaging Subset of DDBSM (CBIS-DDSM)~\cite{lee2017curated}, and the DenseNet-121 neural network for the binary classification task of healthy or tuberculosis cases in the CheXpert Dataset that contains chest X-ray images.}

\renata{The results were compared with GradCAM and Saliency and the authors concluded that their framework was able to provide better accuracy in map the pathology tissue when compared with the ground truth of the images. Furthermore, they described that their framework outperforms the time expended to derive the maps in the mammography and CheXpert datasets when compared to the other two known explainable artificial intelligence methods.}%

\subsection{X-MIR}%

\renata{The paper~\cite{hu2022x} presents an alternative solution to evaluate and improve three different types of similarity-based saliency maps using medical image retrieval and qualitatively computing the insertion and deletion of image regions from the generated saliency maps. The datasets used in the papers were COVID-19 chest X-ray ~\cite{wang2020covid} and International Skin Imaging Collaboration 2017 (ISIC) ~\cite{8363547}. The images from COVID-19 chest X-ray were labeled as normal, pneumonia, and COVID-19 cases. This dataset is highly unbalanced, having only 500 COVID-19 images in the 14.000 from the training dataset. The International Skin Imaging Collaboration (ISIC) dataset contains images that belong to one of three categories: benign nevi, seborrheic keratosis, and melanoma. It also is highly unbalanced with fewer cases of keratosis and melanoma. To create a balanced test set, the authors randomly subselected 90 examples of nevi and melanoma to match the number of available keratosis test examples. Both datasets were classified using the pretrained DenseNet-121 architecture.}

\renata{The main contributions of the paper were: (1) Developed a deep metric learning framework to find similar images from the query images and the public dataset. This framework is a supervised learning approach where the model learns to group images of the same class closer together in a low-dimensional latent space and further away from images there's not of the same class. (2) Apply similarity-based saliency maps to the medical imaging domain, providing visual explanations for deep metric learning models trained on medical images. (3) Adapt a set of metrics to qualitatively measure the similarity-based saliency maps using the insertion and deletion metrics based on the cosine similarity between feature embeddings. (4) Show a form of self-similarity and differential saliency which can highlight regions of images responsible for different disease conditions ~\cite{hu2022x}. As a conclusion, the authors showed that the different forms of saliency maps and the metrics used for quantifying similarity-based saliency maps in image retrieval are promising XAI techniques for medical imaging domain.} %

\subsection{Artificial intelligence: Deep learning in oncological radiomics and
challenges of interpretability and data harmonization}%

\vilaca{The paper~\cite{papadimitroulas2021artificial} explains the possible impacts of Machine Learning (ML)-based Artificial Intelligence (AI) in Healthcare, specifically in Oncology, due to cancer being a worldwide issue with expected growth in the number of deaths ~\cite{bray2018global}. Therefore, the paper reviews the basics of radiomics feature extraction, Deep Convolutional Neural Networks (CNNs) state-of-the-art performances in image processing (including medical images), and interpretability methods that help enable explainable AI (XAI). The amount of available data is usually sufficient for training deep networks, leading to state-of-the-art performances in image processing using CNNs ~\cite{litjens2017survey}. However, there is a comparatively smaller amount of available training samples where labels are on a patient basis (versus millions of images in ImageNet used in CNNs, to attempt predictive modeling in radiomics, leading to no large improvements compared to the standard feature-based radiomics approach~\cite{russakovsky2015imagenet}. To explain that, the authors present how deep learning radiomics (DLR) feature works. In brief, the effectiveness of DLR features is highly related to the quality of the segmentation and the volume of the training dataset~\cite{avanzo2020machine}, in contrast to feature-based radiomics, causing this limitation. While this may be true, the radiomics community relies on Deep Learning (DL) techniques to address challenges found in the standard approach~\cite{hatt2019machine}, like automation on the detection and segmentation steps, harmonizing images, and achieving predictions, even though that address new challenges and issues, like the need for appropriate training with data augmentation techniques, constraints and prior knowledge due to the limited size of available datasets and their high level of heterogeneity, especially when training networks from scratch and the lack of interpretability of these models.

The paper explains that deep learning models are usually referred to as "black box" algorithms, where the final outputs are accepted without justifications. With this in mind, XAI is introduced, as it can bring several benefits for radiomic models relying on deep learning methods. Understanding how the models learn data and arrive at their predictions would allow specialists to improve them, increase their confidence in relying on them and consequently increase the patient's confidence in the tools used, making XAI a crucial criterion for widespread adoption of radiomics. Thus, the authors explain three categories of interpretability methods used in deep learning models, by generally resuming them and their main studies so far. Local Interpretable Model-agnostic Explanations (LIME) can explain a complex non-linear model like DNNs by fitting a locally linear model for a certain prediction~\cite{ribeiro2016should}, visualization of intermediate features to identify which features have actually been learned by convolutional layers of CNNs, and importance estimator to estimate which input pixels are most relevant for specific predictions.

At last, the paper approaches methods for imaging data processing using AI, considering challenges like de-identification of patient-sensitive data for usage, labeling datasets for supervised training, and worries about overfitting performances. The authors explain how multi-center recruitment and harmonization procedures of images and/or radiomic features are needed to establish potential clinical value for radiomics, since most developed models use small internal datasets, accomplishing weak performances compared to clinical standard variables. Two approaches are being studied to allow multi-center-harmonization of datasets to overcome its implications.
The first one is harmonizing in the image domain, by standardization of images procedures or processing images. Only one recent study~\cite{li2021normalization} was able to evaluate this harmonization approach, by showing that relying on harmonized images to extract radiomic features improves the performance of a lasso classifier by an average of 11\%.
The second approach is harmonization in the feature domain, consisting in harmonizing the values of the features. The most promising method for this approach was Combat (combatting batch effect), designed to estimate a batch-specific transformation to express all data in a common space devoid of center effects~\cite{johnson2007adjusting}.
The results from the studies evaluated were not deeply explained.
The paper~\cite{papadimitroulas2021artificial} was very well written and was proven a solid study to clarify the current limitations, challenges, and possible solutions for using ML in the radiomics field for clinical practice.}

\subsection{A Convolutional Neural Network-based Mobile Application to Bedside Neonatal Pain Assessment}%

\nayara{In this paper \cite{painassessement2021} the authors present a framework to identify pain in the facial expression of neonates. The framework is based on CNN (Convolutional Neural Network). Additionally the authors provide access to the framework through a mobile app. To do so they first apply a face localization algorithm called Retina Face to recognize faces on images. Additionally, they used TensorFlow to generate augmented images to increase the data available for the model. They also used TensorFlow for training and testing the CNN.

The classification model was based on a VGG-Face architecture with 16 layers using the transfer-learning method by adding a classifier on top that was trained with neonatal faces captured before and after a painful clinical procedure so they could classify it as pain or no pain. The training proceeded with three classification models where the first used the UNIFESP dataset only, the second one used the ICOPE dataset and the third used both. Data showed that the third model presented better performance than the other two, although its performance was just slightly better than the model with only ICOPE. This performance was considered suited for neonatal pain assessment because it achieved similar results as commonly used clinical pain scales.

To understand the classification of the AI, the authors used the Integrated Gradient method. Through this method it is possible to attribute the importance of an image area for the AI. As a result, the authors mentioned the most important areas of the image for the AI decision would be: the nasolabial groove, the open mouth and the protusion of the tongue. These areas also agree with the visual perception of adults when perceiving pain and are deemed clinically relevant.
The main contribution of the article is the framework of neonatal pain classification, but it is also important to notice the app mobile where the framework can be applied as it facilitates the access on the practical field. As for explainable AI, it is good to notice that the relevant regions found by the method coincide with the clinical and even human expectation
}

\subsection{Improving Interpretability in Machine Diagnosis}

\akemi{	The study developed a CNN deep learning model called Med-XAI-Net (explainable artificial intelligence for medical images analysis), to perform automated detection of Geographic Atrophy (GA) presence or absence from OCT volume scans and to provide interpretability by demonstrating which regions of which B-scans show GA, using the Age-Related Eye Disease Study 2 (AREDS2) Ancillary SD OCT Study dataset.

	Med-XAI-Net simulates the human diagnostic process by using a backbone network consisting of multiple convolutional layers to extract powerful feature representations from each B-scan, followed by a region-attention module to locate the most relevant region in each B-scan, and a image-attention module to select the most relevant B-scans for classifying GA presence or absence in each OCT volume scan.
	
	To evaluate the performance of Med-XAI-Net on classifying GA presence or absence on SD OCT volume scans, its performance was compared with that of popular methods: Baseline, Inflated 3D Convnet (I3D)~\cite{pmlr-v37-ioffe15}, and AttentionNet~\cite{ilse2018attention}. Med-XAI-Net achieved the highest overall accuracy area under the ROC curve , F1 score, and specificity among the 4 models. Its sensitivity was intermediate between that of AttentionNet and Baseline.
	
	To assess the performance of Med-XAI-Net in providing interpretability in selecting the relevant B-scans from a volume scan and localizing GA within a B-scan, three ophthalmologists independently evaluated these selected B-scans and regions without access to any additional clinical information. The performance of the ophthalmologists was essentially identical when using either all B-scans or only the 5 B-scans selected by Med-XAI-Net as most representative, but it was lower (driven by lower sensitivity) when using only the 1 region of 1 B-scan selected by Med-XAI-Net compared with using all B-scans or the 5 B-scans.}

\subsection{ExAID: A Multimodal Explanation Framework for Computer-Aided Diagnosis of Skin Lesions}

\fabiana{The paper\cite{lucieri2022exaid} explains the functionality of the framework ExAID and its use to identify skin lesions. The framework is based in the Concept Activation Vector (CAVs) to map the human-understandable concepts, and Concept Location Maps(CLMs) to highlight concepts in the input space.
	The framework also provides tools for medical and educational researches, establishing a transparent and understandable integration of AIs into medical workflow.
	
	The training set for disease level classification consists of Melanoma and Nevi images obtained from ISIC 2019~\cite{8363547, combalia2019bcn20000, tschandl2018ham10000}, PH2~\cite{6610779} and derm7pt~\cite{8333693} datasets. The ISIC 2019 dataset is a public collection with 25,331 images of different provenances divided into wight different classes.
	
	For the training of classifiers specific dermoscopic concepts are needed, which normally is not available in these datasets, limiting our train selection and assessment specially in the PH2 and derm7pt datasets. These datasets provide color and lesion segmentation masks, and extensive well-selected annotations regarding the presence or absence of various concepts. ExAID is a generic toolbox for human-centered explanations based in Deep Learning models, even beyond dermatologic context. In addiction to the Deep Learning model, ExAID is composed of 3 basic components: the Concept of Identification, Concept Location, and Decision Explanation modules.
	
	The results are coherent texts and a simple explanation, giving evaluations with absence, moderate evidence and strong evidence of concepts.
	
	With ExAID, it is possible to view the location of concepts simultaneously for all samples in a dataset. This allows for a quick examination of the localization behavior of a model’s concept, helping to validate the behavior of the system, and identify possible systematic errors in the dataset, thus revealing patterns in the localization process.

}

\subsection{ExplAIn: Explanatory Artificial Intelligence for Diabetic Retinopathy Diagnosis }

\fabiana{Diabetic Retinopathy (DR) is a leading and growing cause of visual impairment and blindness: by 2040, about 600 million people worldwide will have diabetes~\cite{ogurtsova2017idf}, a third of whom will have DR~\cite{yu2021reconfigisp}. %
	
	To improve RD screening programs, numerous Artificial Intelligence (AI) programs have thus been developed to automate RD diagnosis using CFP~\cite{ting2019deep}. However, due to the “black box” that turned out to be the nature of next-gen AI, these systems have yet to earn the trust of doctors and patients.
	
	In this scenario, eXplanatory Artificial Intelligence (XAI) appears to gain confidence, and bring a very desirable explanation. This algorithm learns to segment and categorize lesions in images; the final image-level classification derives directly from these multivariate lesion segments. The novelty of this explanatory framework is that it is trained from end to end, with only image supervision, the injury concepts and their categories emerging by themselves. The advantage of such an architecture is that automatic diagnoses can be explained simply by images and/or a few sentences. ExplAIn is evaluated at the image level and the pixel level in various CFP image datasets.
	
	Unlike previous visualization methods, ExplAIn does not attempt to retrospectively analyze a complex classification process. Instead, it modifies the classification process in such a way that it can be understood directly. Explanatory Artificial Intelligence (XAI) is a growing field of research~\cite{gilpin2018explaining} motivated by security-conscious potential users of AI~\cite{russell2015research}. It is also driven by European regulations and others, its purpose being granting users the right to an explanation about algorithms, and decisions that have been made about them~\cite{goodman2017european}. The authors hope that this new framework, which offers a high classification and explainability performance, will facilitate the implementation of AI and gain the trust of physicians and patients.
	
}

\subsection{COVID-19 Automatic Diagnosis with Radiographic Imaging: Explainable Attention Transfer Deep Neural Networks}

\fabiana{During the COVID-19 pandemic, caused by the same Severe Acute Respiratory Syndrome Coronavirus 2 (SARS-CoV-2), researchers are seeking help of deep learning as a method to alleviate the enormous burden of reading radiological images by clinicians. Currently, Reverse Transcription Polymerase Chain Reaction (RT-PCR) is the universally applicable and effective method for diagnosing COVID-19. However, there is a conflict between the scarcity of equipment for testing environments and rapid and accurate screening of suspicious individuals. Furthermore, the diagnosis of infection by clinicians and radiologists is a highly subjective task, often influenced by individual biases and clinical experience.
	
	For this reason, the use of Artificial Intelligence comes to help diagnoses to be more reliable. To automatically differentiate COVID-19 and community-acquired pneumonia from healthy lungs on radiographic images, an explainable care transfer classification model based on the structure of the knowledge distillation network is proposed. The attention transfer direction always goes from the teacher network to the student network. First, the teacher network extracts global resources and focuses on infecting regions to generate maps of care. Also, a image fusion module is used to combine attention knowledge transferred from the teacher network to the student network with the essential information in the original input. While the teacher network focuses on global features, the student branch focuses on regions of irregularly shaped lesions to learn discriminative features. For data collection, we performed extensive experiments on public chests and X-ray and CT datasets to demonstrate the explainability of the proposed architecture in the diagnosis of COVID-19.
	
	The main contribution of this work is an explainable model of attention transfer classification based on the structure of the knowledge distillation network that is designed to achieve an automatic diagnosis of COVID-19 with radiology.	A deformable attention module is proposed to focus on irregularly shaped regions of infection and their surroundings on radiological images. Extensive publicly available experiments were performed with chest radiography and CT datasets to evaluate the proposed multiclass classification model, differentiating COVID-19, normal, and CAP cases. In addition, the algorithm achieves state-of-the-art performance and improves model explainability by salience map, severity assessment, and prediction confidence.
	
}

\subsection{Improving Uncertainty Estimation With Semi-Supervised Deep Learning for Covid-19 Detection Using Chest X-Ray Images}%

\jose{This paper~\cite{calderon2021improving} is developed with the premise that x-ray chest is, in general, more widely accessible when compared to computed tomography imaging.
	
To improve the reliability of uncertainty estimations, the authors used MixMatch~\cite{mixmatch}, which creates a set of pseudo-labels, also implements an unsupervised regularization term and tested three different uncertainty estimation methods Softmax, Monte Carlo Dropout (MCD)~\cite{gal2016dropout} and Deterministic Uncertainty Quantification (DUQ)~\cite{van2020simple}. To compare the evaluated methods, the Jensen-Shannon (JS) distance is used. 
	
Also, in the architecture, the authors used WideResNet model in supervised and semi-supervised models. The conclusion obtained is that with a low number of labels, the JS divergence is boosted, and their use is recommended.

Moreover, the authors affirms ‘Epistemic uncertainty can be considered to be very high in models trained with very few labels, as the feature space sample is very limited’.
	
The data sample of COVID-19+ was downloaded from the publicly available GitHub repository of Cohen, and to do their tests, both models are trained with nl = 20~100 labelled observations, being divided with 198 observations, 70\% (138 observations) to train and 30\% (60 observations) to test. The authors comment about the low availability of public repositories of COVID-19 chest X-rays with labels creates a limitation to them.
}

\newpage

\section{Results}
\label{results}
From the previous analysis, we were able to extract important information common to all papers and understand some shortcomings of XAI in medicine, as well as detect emerging trends in this field. A quantitative analysis is performed on the models used, explainable frameworks applied and pathologies targeted, all presented in the tables below.

\renewcommand{\arraystretch}{1.5}
\noindent\begin{minipage}[t]{0.5\textwidth}%
	\begin{center}
		\captionof{table}{XAI methodologies}
		\begin{tabular}{p{5cm}|p{1cm}}
			\hline
			\multicolumn{2}{c}{Studies that applied XAI on Medical Images} \\
			\hline
			
			Explainable Models & {\itshape Quant.} \\ [1ex] 
			
			LIME & 7  \\ 
			
			SHAP & 6  \\
			
			Grad-CAM & 4  \\
			
			CAV & 3 \\
			
			CLM & 2 \\ 
			
			Salience Map & 2 \\ 
			
			Guided Backpropagation & 2 \\ 
			
			XEGK & 1 \\ 
			
			Guided Grad-CAM & 1 \\
			
			Gradient Attention Rollout & 1 \\
			
			DeepLIFT & 1 \\
			
			t-SNE & 1 \\		[1ex] 
			\hline
			
		\end{tabular}
	\end{center}
\end{minipage}
\begin{minipage}[t]{0.5\textwidth}%
	\begin{center}
		\captionof{table}{Architectures applied}
		\begin{tabular}{p{3cm}|p{1cm}}
			\hline
			\multicolumn{2}{c}{Studies that trained CNN models} \\
			\hline
			
			Models & {\itshape Quant.} \\ [1ex] 
			
			SqueezeNet & 3  \\ 
			
			MobileNet & 2  \\
			
			WideResNet & 2  \\
			
			VGG-16 & 2 \\
			
			EfficientNet & 1 \\ 
			
			VGG-Face & 1 \\ 
			
			ResNet50 & 1 \\ 
			
			Inception v3 & 1 \\

			DenseNet-121 & 1  \\ 
			
			DenseNet-169 & 1  \\ 

			USRNet & 1  \\
			
			U-net & 1 \\ [1ex] 
			\hline
			\multicolumn{2}{c}{Studies that trained other models} \\
			\hline

			Vision Transformer & 1 \\

			\hline
		\end{tabular}
		
	\end{center}
\end{minipage}
\break

\begin{center}
	\captionof{table}{Medical images domain}
	\begin{tikzpicture}[scale=0.8]
		\pie[color = {red!80,red!60!orange!80,orange!90, yellow!80,yellow!70!green!90,green!60, green!70!blue, cyan!60!green!90!, blue!60, purple!40!blue!50, pink!60!purple!90},
		sum = auto,
		text = legend
		]{8/Pulmonary Diseases,
			3/Skin Lesion,
			2/Facial Expression,
			1/Malaria Cells,
			1/Brain Tumors,
			1/Mitosis,
			1/Parkinson's Disease,
			1/Bladder Cancer,
			1/Neonate Thermograms,
			1/Geographic Atrophy,
			1/Diabetic Retinopathy}
	\end{tikzpicture}
\end{center}

\subsection{Discussion}
Complementing the quantitative analysis, we evaluate qualitatively the works presented and highlight outstanding issues within the field. Most notably, we bring light to the following points:
\begin{itemize}
	\item A standard metric or procedure to evaluate the effectiveness of XAI methods does not exist. This arises mainly as a consequence of the meaning of "interpretability" and its varying definitions for every applicable scenario, leading every author to evaluate the explainability of their model in their own manner.
	\item Similarly, no such standardization for medical images and models exist either. Because medicine is an extremely rich field, to the best of our knowledge there is no simple way to globally streamline the performance evaluation of every application.
	\item As a result of the modality of present works, CNNs are widely applied. However, these architectures usually require great amounts of data to properly learn powerful characteristics~\cite{papadimitroulas2021artificial}. And since the field of medicine, to the best of our knowledge, does not contain a common, expansive and condensed database for such applications, every work has to spend valuable time gathering data.
	\item Many datasets used in the reviewed works are private, a consequence of the secrecy required in many medical fields. This renders their contributions harder to reproduce.
	\item Some works, instead of developing an automatic XAI system, utilize experts to analyze their models predictions.
\end{itemize}

\section{Conclusion}
\label{s.conclusion}

\jose{Although several AI models reach acceptable conclusions and with high accuracy, most of them have a reliability-related deficit due to the lack of transparency of models, turning them into a ‘black-box’ solution. However, XAI comes to contour this problem, providing a better look about the influence of each input data. In this work, the explicability study method focused on the medicinal field, demonstrating the interest of professionals in models that can evaluate and bring new opinions to the clinic. This is possible thanks to the use of Deep Learning, which in general, demonstrates many advances to recognize and evaluate medical images. In addition, this paper introduces basic concepts about XAI, with evaluation of most recent works, and an overview of the field. The 32 works surveyed in this paper were selected based on their novelty, importance in medicine and the applied explainable deep learning techniques. Some use well-established eXplainable AI techniques, while others focus on experts' analysis for understanding their models' classifications. Nonetheless, while XAI still stands as a rich and varied field, many trends are emerging at the intersection between medicine and deep learning.}

During the evaluation of the reviewed works, crucial points were detected among them:
\begin{itemize}
	\item LIME and SHAP are the most present frameworks in the papers reviewed. Approximately 40\% of works use them in some sense (7 for LIME, 6 for SHAP), mainly for explaining the impact of different image regions in the final prediction. This trend leads us to understand that general-purpose XAI frameworks will be a common ocurrence in the deep learning field for their simultaneous simplicity and interpretability.
	\item To ensure doctors do not give early diagnosis because of routine biases, many models proposed in this work can grant results that make them rethink about the patient, bringing in a second opinion and breaking this cycle.
	\item CNN was the most used machine learning technique in the papers reviewed, representing approximately 56\% of the trained models. The technique was broadly used due to it being a popular and effective tool for medical image understanding, producing highly accurate recognition results.
	\item Many pathologies and conditions were targeted in the works reviewed - from neonatal thermogram health assessment to eye fundus analysis -, but COVID-19 and pneumonia were tackled the most, representing 25\% of the pathologies discussed. Because our survey focuses on works published since 2020, the year the coronavirus pandemic became global, the medical sciences spent much effort to quickly and effectively diagnose COVID-19 (and separate its occurrences from general pneumonia) in order to mitigate this diseases' damages.
\end{itemize}

To further enrich the field of XAI and deep learning applications in medicine, it highlisghts the creation of a dense and plural common database for medical images from which models can learn important features by pretraining. Although the ImageNet dataset was widely used in the reviewed papers, medical images tend to have intricacies not commonly found in such datasets.

\bibliographystyle{unsrt}  
\bibliography{survey}

\begin{thebibliography}{100}

\bibitem{das2020opportunities}
Arun Das and Paul Rad.
\newblock Opportunities and challenges in explainable artificial intelligence
  (xai): A survey.
\newblock {\em arXiv preprint arXiv:2006.11371}, 2020.

\bibitem{8466590}
Amina Adadi and Mohammed Berrada.
\newblock Peeking inside the black-box: A survey on explainable artificial
  intelligence (xai).
\newblock {\em IEEE Access}, 6:52138--52160, 2018.

\bibitem{doi:10.1073/pnas.1900654116}
W.~James Murdoch, Chandan Singh, Karl Kumbier, Reza Abbasi-Asl, and Bin Yu.
\newblock Definitions, methods, and applications in interpretable machine
  learning.
\newblock {\em Proceedings of the National Academy of Sciences},
  116(44):22071--22080, 2019.

\bibitem{8419428}
Muhammad~Aurangzeb Ahmad, Ankur Teredesai, and Carly Eckert.
\newblock Interpretable machine learning in healthcare.
\newblock In {\em 2018 IEEE International Conference on Healthcare Informatics
  (ICHI)}, pages 447--447, 2018.

\bibitem{VANDERVELDEN2022102470}
Bas~H.M. {van der Velden}, Hugo~J. Kuijf, Kenneth~G.A. Gilhuijs, and Max~A.
  Viergever.
\newblock Explainable artificial intelligence (xai) in deep learning-based
  medical image analysis.
\newblock {\em Medical Image Analysis}, 79:102470, 2022.

\bibitem{mitchel1997machine}
T~Mitchel.
\newblock Machine learning, mcgraw-hill education (ise editions).
\newblock 1997.

\bibitem{TAN06}
Pang-Ning Tan, Michael Steinbach, and Vipin Kumar.
\newblock {\em Introduction to Data Mining}.
\newblock Addison-Wesley, Boston, 2013.

\bibitem{Lipton2016mythos}
Zachary~Chase Lipton.
\newblock The mythos of model interpretability.
\newblock {\em CoRR}, abs/1606.03490, 2016.

\bibitem{Freitas2013comprehensible}
Alex~A. Freitas.
\newblock Comprehensible classification models: A position paper.
\newblock {\em SIGKDD Explor. Newsl.}, 15(1):1–10, mar 2014.

\bibitem{DBLP:journals/corr/Miller17a}
Tim Miller.
\newblock Explanation in artificial intelligence: Insights from the social
  sciences.
\newblock {\em CoRR}, abs/1706.07269, 2017.

\bibitem{shung2012principles}
K~Kirk Shung, Michael~B Smith, and Benjamin~MW Tsui.
\newblock {\em Principles of medical imaging}.
\newblock Academic Press, 2012.

\bibitem{hu2022x}
Brian Hu, Bhavan Vasu, and Anthony Hoogs.
\newblock X-mir: Explainable medical image retrieval.
\newblock In {\em Proceedings of the IEEE/CVF Winter Conference on Applications
  of Computer Vision}, pages 440--450, 2022.

\bibitem{lucieri2022exaid}
Adriano Lucieri, Muhammad~Naseer Bajwa, Stephan~Alexander Braun, Muhammad~Imran
  Malik, Andreas Dengel, and Sheraz Ahmed.
\newblock Exaid: A multimodal explanation framework for computer-aided
  diagnosis of skin lesions, 2022.

\bibitem{stieler2021skinimage}
Fabian Stieler, Fabian Rabe, and Bernhard Bauer.
\newblock Towards domain-specific explainable ai: Model interpretation of a
  skin image classifier using a human approach.
\newblock In {\em 2021 IEEE/CVF Conference on Computer Vision and Pattern
  Recognition Workshops (CVPRW)}, pages 1802--1809, 2021.

\bibitem{lucieri2020interpretability}
Adriano Lucieri, Muhammad~Naseer Bajwa, Stephan~Alexander Braun, Muhammad~Imran
  Malik, Andreas Dengel, and Sheraz Ahmed.
\newblock On interpretability of deep learning based skin lesion classifiers
  using concept activation vectors.
\newblock In {\em 2020 international joint conference on neural networks
  (IJCNN)}, pages 1--10. IEEE, 2020.

\bibitem{lenis2020domain}
Dimitrios Lenis, David Major, Maria Wimmer, Astrid Berg, Gert Sluiter, and
  Katja B{\"u}hler.
\newblock Domain aware medical image classifier interpretation by
  counterfactual impact analysis.
\newblock In {\em International Conference on Medical Image Computing and
  Computer-Assisted Intervention}, pages 315--325. Springer, 2020.

\bibitem{brunese2020explainable}
Luca Brunese, Francesco Mercaldo, Alfonso Reginelli, and Antonella Santone.
\newblock Explainable deep learning for pulmonary disease and coronavirus
  covid-19 detection from x-rays.
\newblock {\em Computer Methods and Programs in Biomedicine}, 196:105608, 2020.

\bibitem{corizzo2021explainable}
Roberto Corizzo, Yohan Dauphin, Colin Bellinger, Eftim Zdravevski, and Nathalie
  Japkowicz.
\newblock Explainable image analysis for decision support in medical
  healthcare.
\newblock In {\em 2021 IEEE International Conference on Big Data (Big Data)},
  pages 4667--4674. IEEE, 2021.

\bibitem{mondal2021xvitcos}
Arnab~Kumar Mondal, Arnab Bhattacharjee, Parag Singla, and AP~Prathosh.
\newblock xvitcos: explainable vision transformer based covid-19 screening
  using radiography.
\newblock {\em IEEE Journal of Translational Engineering in Health and
  Medicine}, 10:1--10, 2021.

\bibitem{bang2021spatio}
Ji-Seon Bang, Min-Ho Lee, Siamac Fazli, Cuntai Guan, and Seong-Whan Lee.
\newblock Spatio-spectral feature representation for motor imagery
  classification using convolutional neural networks.
\newblock {\em IEEE Transactions on Neural Networks and Learning Systems},
  PP:1--12, 01 2021.

\bibitem{li2021explainable}
Jingchen Li, Haobin Shi, and Kao-Shing Hwang.
\newblock An explainable ensemble feedforward method with gaussian
  convolutional filter.
\newblock {\em Knowledge-Based Systems}, 225:107103, 2021.

\bibitem{MOHAGHEGHI2022105106}
Saeed Mohagheghi and Amir~Hossein Foruzan.
\newblock Developing an explainable deep learning boundary correction method by
  incorporating cascaded x-dim models to improve segmentation defects in liver
  ct images.
\newblock {\em Computers in Biology and Medicine}, 140:105106, 2022.

\bibitem{Yang2021review}
Sijie Yang, Fei Zhu, Xinghong Ling, Quan Liu, and Peiyao Zhao.
\newblock Intelligent health care: Applications of deep learning in
  computational medicine.
\newblock {\em Frontiers in Genetics}, 12, 2021.

\bibitem{doi:10.1177/2053951715622512}
Jenna Burrell.
\newblock How the machine ‘thinks’: Understanding opacity in machine
  learning algorithms.
\newblock {\em Big Data \& Society}, 3(1):2053951715622512, 2016.

\bibitem{10.1145/2939672.2945386}
Sara Hajian, Francesco Bonchi, and Carlos Castillo.
\newblock Algorithmic bias: From discrimination discovery to fairness-aware
  data mining.
\newblock In {\em Proceedings of the 22nd ACM SIGKDD International Conference
  on Knowledge Discovery and Data Mining}, KDD '16, page 2125–2126, New York,
  NY, USA, 2016. Association for Computing Machinery.

\bibitem{DBLP:journals/corr/Wang17j}
William~Yang Wang.
\newblock "liar, liar pants on fire": {A} new benchmark dataset for fake news
  detection.
\newblock {\em CoRR}, abs/1705.00648, 2017.

\bibitem{SHI2021100038}
Xiaoshuang Shi, Tiarnan~D.L. Keenan, Qingyu Chen, Tharindu {De Silva}, Alisa~T.
  Thavikulwat, Geoffrey Broadhead, Sanjeeb Bhandari, Catherine Cukras, Emily~Y.
  Chew, and Zhiyong Lu.
\newblock Improving interpretability in machine diagnosis: Detection of
  geographic atrophy in oct scans.
\newblock {\em Ophthalmology Science}, 1(3):100038, 2021.

\bibitem{QUELLEC2021102118}
Gwenolé Quellec, Hassan {Al Hajj}, Mathieu Lamard, Pierre-Henri Conze, Pascale
  Massin, and Béatrice Cochener.
\newblock Explain: Explanatory artificial intelligence for diabetic retinopathy
  diagnosis.
\newblock {\em Medical Image Analysis}, 72:102118, 2021.

\bibitem{shi2021covid}
Wenqi Shi, Li~Tong, Yuanda Zhu, and May~D Wang.
\newblock Covid-19 automatic diagnosis with radiographic imaging: Explainable
  attention transfer deep neural networks.
\newblock {\em IEEE Journal of Biomedical and Health Informatics},
  25(7):2376--2387, 2021.

\bibitem{hossain2020covid19}
M.~Shamim Hossain, Ghulam Muhammad, and Nadra Guizani.
\newblock Explainable ai and mass surveillance system-based healthcare
  framework to combat covid-i9 like pandemics.
\newblock {\em IEEE Network}, 34(4):126--132, 2020.

\bibitem{calderon2021improving}
Saul Calderon-Ramirez, Shengxiang Yang, Armaghan Moemeni, Simon
  Colreavy-Donnelly, David~A Elizondo, Luis Oala, Jorge
  Rodr{\'\i}guez-Capit{\'a}n, Manuel Jim{\'e}nez-Navarro, Ezequiel
  L{\'o}pez-Rubio, and Miguel~A Molina-Cabello.
\newblock Improving uncertainty estimation with semi-supervised deep learning
  for covid-19 detection using chest x-ray images.
\newblock {\em Ieee Access}, 9:85442--85454, 2021.

\bibitem{ong2021comparative}
Joe~Huei Ong, Kam~Meng Goh, and Li~Li Lim.
\newblock Comparative analysis of explainable artificial intelligence for
  covid-19 diagnosis on cxr image.
\newblock In {\em 2021 IEEE International Conference on Signal and Image
  Processing Applications (ICSIPA)}, pages 185--190. IEEE, 2021.

\bibitem{wang2021xai}
Zihao Wang, Hang Zhu, Yingnan Ma, and Anup Basu.
\newblock Xai feature detector for ultrasound feature matching.
\newblock In {\em 2021 43rd Annual International Conference of the IEEE
  Engineering in Medicine \& Biology Society (EMBC)}, pages 2928--2931. IEEE,
  2021.

\bibitem{deramgozin2021hybrid}
M~Deramgozin, Slavisa Jovanovic, Hassan Rabah, and Naeem Ramzan.
\newblock A hybrid explainable ai framework applied to global and local facial
  expression recognition.
\newblock In {\em 2021 IEEE International Conference on Imaging Systems and
  Techniques (IST)}, pages 1--5. IEEE, 2021.

\bibitem{raihan2022malaria}
Md~Raihan, Abdullah-Al Nahid, et~al.
\newblock Malaria cell image classification by explainable artificial
  intelligence.
\newblock {\em Health and Technology}, 12(1):47--58, 2022.

\bibitem{pianpanit2021parkinson}
Theerasarn Pianpanit, Sermkiat Lolak, Phattarapong Sawangjai, Thapanun
  Sudhawiyangkul, and Theerawit Wilaiprasitporn.
\newblock Parkinson’s disease recognition using spect image and interpretable
  ai: A tutorial.
\newblock {\em IEEE Sensors Journal}, 21(20):22304--22316, 2021.

\bibitem{uehara2021explainable}
Kazuki Uehara, Masahiro Murakawa, Hirokazu Nosato, and Hidenori Sakanashi.
\newblock Explainable feature embedding using convolutional neural networks for
  pathological image analysis.
\newblock In {\em 2020 25th International Conference on Pattern Recognition
  (ICPR)}, pages 4560--4565. IEEE, 2021.

\bibitem{o2022explainable}
S~O’Sullivan, M~Janssen, Andreas Holzinger, Nathalie Nevejans, O~Eminaga,
  CP~Meyer, and Arkadiusz Miernik.
\newblock Explainable artificial intelligence (xai): closing the gap between
  image analysis and navigation in complex invasive diagnostic procedures.
\newblock {\em World Journal of Urology}, pages 1--10, 2022.

\bibitem{painassessement2021}
Lucas~P Carlini, Leonardo~A Ferreira, Gabriel~AS Coutrin, Victor~V Varoto,
  Tatiany~M Heiderich, Rita~CX Balda, Marina~CM Barros, Ruth Guinsburg, and
  Carlos~E Thomaz.
\newblock A convolutional neural network-based mobile application to bedside
  neonatal pain assessment.
\newblock In {\em 2021 34th SIBGRAPI Conference on Graphics, Patterns and
  Images (SIBGRAPI)}, pages 394--401. IEEE, 2021.

\bibitem{li2020neural}
Xiang Li and Shihao Ji.
\newblock Neural image compression and explanation.
\newblock {\em IEEE Access}, 8:214605--214615, 2020.

\bibitem{rasaee2021modresnet}
Hamza Rasaee and Hassan Rivaz.
\newblock Explainable ai and susceptibility to adversarial attacks: a case
  study in classification of breast ultrasound images, 2021.

\bibitem{papadimitroulas2021artificial}
Panagiotis Papadimitroulas, Lennart Brocki, Neo~Christopher Chung, Wistan
  Marchadour, Franck Vermet, Laurent Gaubert, Vasilis Eleftheriadis, Dimitris
  Plachouris, Dimitris Visvikis, George~C Kagadis, et~al.
\newblock Artificial intelligence: Deep learning in oncological radiomics and
  challenges of interpretability and data harmonization.
\newblock {\em Physica Medica}, 83:108--121, 2021.

\bibitem{arefin2021non}
Raisul Arefin, Manar~D Samad, Furkan~A Akyelken, and Arash Davanian.
\newblock Non-transfer deep learning of optical coherence tomography for
  post-hoc explanation of macular disease classification.
\newblock In {\em 2021 IEEE 9th International Conference on Healthcare
  Informatics (ICHI)}, pages 48--52. IEEE, 2021.

\bibitem{civit2020dual}
Javier Civit-Masot, Manuel~J Dom{\'\i}nguez-Morales, Saturnino
  Vicente-D{\'\i}az, and Anton Civit.
\newblock Dual machine-learning system to aid glaucoma diagnosis using disc and
  cup feature extraction.
\newblock {\em IEEE Access}, 8:127519--127529, 2020.

\bibitem{kabakcci2021automated}
Kaan~Aykut Kabak{\c{c}}{\i}, Asl{\i} {\c{C}}ak{\i}r, {\.I}lknur T{\"u}rkmen,
  Beh{\c{c}}et~U{\u{g}}ur T{\"o}reyin, and Abdulkerim {\c{C}}apar.
\newblock Automated scoring of cerbb2/her2 receptors using histogram based
  analysis of immunohistochemistry breast cancer tissue images.
\newblock {\em Biomedical Signal Processing and Control}, 69:102924, 2021.

\bibitem{liz2021ensembles}
Helena Liz, Manuel S{\'a}nchez-Monta{\~n}{\'e}s, Alfredo Tagarro, Sara
  Dom{\'\i}nguez-Rodr{\'\i}guez, Ron Dagan, and David Camacho.
\newblock Ensembles of convolutional neural network models for pediatric
  pneumonia diagnosis.
\newblock {\em Future Generation Computer Systems}, 122:220--233, 2021.

\bibitem{turkish9302311}
Ahmet~Haydar Örnek and Murat Ceylan.
\newblock Explainable features in classification of neonatal thermograms.
\newblock In {\em 2020 28th Signal Processing and Communications Applications
  Conference (SIU)}, pages 1--4, 2020.

\bibitem{molnar2022}
Christoph Molnar.
\newblock {\em Interpretable Machine Learning}.
\newblock 2 edition, 2022.

\bibitem{murdoch2019definitions}
W~James Murdoch, Chandan Singh, Karl Kumbier, Reza Abbasi-Asl, and Bin Yu.
\newblock Definitions, methods, and applications in interpretable machine
  learning.
\newblock {\em Proceedings of the National Academy of Sciences},
  116(44):22071--22080, 2019.

\bibitem{selvaraju2017grad}
Ramprasaath~R Selvaraju, Michael Cogswell, Abhishek Das, Ramakrishna Vedantam,
  Devi Parikh, and Dhruv Batra.
\newblock Grad-cam: Visual explanations from deep networks via gradient-based
  localization.
\newblock In {\em Proceedings of the IEEE international conference on computer
  vision}, pages 618--626, 2017.

\bibitem{kim2017interpretability}
Been Kim, Martin Wattenberg, Justin Gilmer, Carrie Cai, James Wexler, Fernanda
  Viegas, and Rory Sayres.
\newblock Interpretability beyond feature attribution: Quantitative testing
  with concept activation vectors (tcav).
\newblock 2017.

\bibitem{molnar2020interpretable}
Christoph Molnar.
\newblock {\em Interpretable machine learning}.
\newblock Lulu. com, 2020.

\bibitem{shrikumar2017learning}
Avanti Shrikumar, Peyton Greenside, and Anshul Kundaje.
\newblock Learning important features through propagating activation
  differences.
\newblock In {\em International conference on machine learning}, pages
  3145--3153. PMLR, 2017.

\bibitem{simonyan2013deep}
Karen Simonyan, Andrea Vedaldi, and Andrew Zisserman.
\newblock Deep inside convolutional networks: Visualising image classification
  models and saliency maps.
\newblock 2013.

\bibitem{Guided_Backpropagation}
Jost Springenberg, Alexey Dosovitskiy, Thomas Brox, and Martin Riedmiller.
\newblock Striving for simplicity: The all convolutional net.
\newblock 12 2014.

\bibitem{10.1371/journal.pone.0130140}
Sebastian Bach, Alexander Binder, Grégoire Montavon, Frederick Klauschen,
  Klaus-Robert Müller, and Wojciech Samek.
\newblock On pixel-wise explanations for non-linear classifier decisions by
  layer-wise relevance propagation.
\newblock {\em PLOS ONE}, 10(7):1--46, 07 2015.

\bibitem{DBLP:journals/corr/abs-2009-07896}
Narine Kokhlikyan, Vivek Miglani, Miguel Martin, Edward Wang, Bilal Alsallakh,
  Jonathan Reynolds, Alexander Melnikov, Natalia Kliushkina, Carlos Araya, Siqi
  Yan, and Orion Reblitz{-}Richardson.
\newblock Captum: {A} unified and generic model interpretability library for
  pytorch.
\newblock {\em CoRR}, abs/2009.07896, 2020.

\bibitem{NIPS2017_8a20a862}
Scott~M Lundberg and Su-In Lee.
\newblock A unified approach to interpreting model predictions.
\newblock In I.~Guyon, U.~Von Luxburg, S.~Bengio, H.~Wallach, R.~Fergus,
  S.~Vishwanathan, and R.~Garnett, editors, {\em Advances in Neural Information
  Processing Systems}, volume~30. Curran Associates, Inc., 2017.

\bibitem{DBLP:journals/corr/RibeiroSG16}
Marco~T{\'{u}}lio Ribeiro, Sameer Singh, and Carlos Guestrin.
\newblock "why should {I} trust you?": Explaining the predictions of any
  classifier.
\newblock {\em CoRR}, abs/1602.04938, 2016.

\bibitem{bankman2008handbook}
Isaac Bankman.
\newblock {\em Handbook of medical image processing and analysis}.
\newblock Elsevier, 2008.

\bibitem{russakovsky2015imagenet}
Olga Russakovsky, Jia Deng, Hao Su, Jonathan Krause, Sanjeev Satheesh, Sean Ma,
  Zhiheng Huang, Andrej Karpathy, Aditya Khosla, Michael Bernstein, et~al.
\newblock Imagenet large scale visual recognition challenge.
\newblock {\em International journal of computer vision}, 115(3):211--252,
  2015.

\bibitem{tjoa2020survey}
Erico Tjoa and Cuntai Guan.
\newblock A survey on explainable artificial intelligence (xai): Toward medical
  xai.
\newblock {\em IEEE transactions on neural networks and learning systems},
  32(11):4793--4813, 2020.

\bibitem{holzinger2019causability}
Andreas Holzinger, Georg Langs, Helmut Denk, Kurt Zatloukal, and Heimo
  M{\"u}ller.
\newblock Causability and explainability of artificial intelligence in
  medicine.
\newblock {\em Wiley Interdisciplinary Reviews: Data Mining and Knowledge
  Discovery}, 9(4):e1312, 2019.

\bibitem{Selvaraju2017gradcam}
Ramprasaath~R. Selvaraju, Michael Cogswell, Abhishek Das, Ramakrishna Vedantam,
  Devi Parikh, and Dhruv Batra.
\newblock Grad-{CAM}: Visual explanations from deep networks via gradient-based
  localization.
\newblock {\em International Journal of Computer Vision}, 128(2):336--359, oct
  2019.

\bibitem{Goodfellow2015attacks}
Ian~J. Goodfellow, Jonathon Shlens, and Christian Szegedy.
\newblock Explaining and harnessing adversarial examples, 2014.

\bibitem{Dhabyani2020dataset}
Walid Al-Dhabyani, Mohammed Gomaa, Hussien Khaled, and Aly Fahmy.
\newblock Dataset of breast ultrasound images.
\newblock {\em Data in Brief}, 28:104863, 2020.

\bibitem{Rodrigues2017dataset}
Paulo~Sergio Rodrigues.
\newblock Breast ultrasound images.
\newblock 2017.

\bibitem{kerasvis}
Raghavendra Kotikalapudi and contributors.
\newblock keras-vis.
\newblock \url{https://github.com/raghakot/keras-vis}, 2017.

\bibitem{kermany2018identifying}
Daniel~S Kermany, Michael Goldbaum, Wenjia Cai, Carolina~CS Valentim, Huiying
  Liang, Sally~L Baxter, Alex McKeown, Ge~Yang, Xiaokang Wu, Fangbing Yan,
  et~al.
\newblock Identifying medical diagnoses and treatable diseases by image-based
  deep learning.
\newblock {\em Cell}, 172(5):1122--1131, 2018.

\bibitem{rimone}
Francisco Fumero, Jose Sigut, M~Alayón, Silvia andGonzález-Hernández, and
  M~González de~la Rosa.
\newblock Interactive tool and database for optic disc and cupsegmentation of
  stereo and monocular retinal fundus images.
\newblock 06 2015.

\bibitem{drishti}
J.~{Sivaswamy}, S.~R. {Krishnadas}, G.~{Datt Joshi}, M.~{Jain}, and A.~U. {Syed
  Tabish}.
\newblock Drishti-gs: Retinal image dataset for optic nerve head(onh)
  segmentation.
\newblock In {\em 2014 IEEE 11th International Symposium on Biomedical Imaging
  (ISBI)}, pages 53--56, April 2014.

\bibitem{qaiser2018her2}
Talha Qaiser, Abhik Mukherjee, Chaitanya Reddy~Pb, Sai~D Munugoti, Vamsi
  Tallam, Tomi Pitk{\"a}aho, Taina Lehtim{\"a}ki, Thomas Naughton, Matt
  Berseth, An{\'\i}bal Pedraza, et~al.
\newblock Her 2 challenge contest: a detailed assessment of automated her 2
  scoring algorithms in whole slide images of breast cancer tissues.
\newblock {\em Histopathology}, 72(2):227--238, 2018.

\bibitem{ozturk2020automated}
Tulin Ozturk, Muhammed Talo, Eylul~Azra Yildirim, Ulas~Baran Baloglu, Ozal
  Yildirim, and U~Rajendra Acharya.
\newblock Automated detection of covid-19 cases using deep neural networks with
  x-ray images.
\newblock {\em Computers in biology and medicine}, 121:103792, 2020.

\bibitem{lime}
Marco~Tulio Ribeiro, Sameer Singh, and Carlos Guestrin.
\newblock "why should {I} trust you?": Explaining the predictions of any
  classifier.
\newblock In {\em Proceedings of the 22nd {ACM} {SIGKDD} International
  Conference on Knowledge Discovery and Data Mining, San Francisco, CA, USA,
  August 13-17, 2016}, pages 1135--1144, 2016.

\bibitem{resnet50}
Kaiming He, Xiangyu Zhang, Shaoqing Ren, and Jian Sun.
\newblock Deep residual learning for image recognition, 2015.

\bibitem{inceptionv3}
Christian Szegedy, Vincent Vanhoucke, Sergey Ioffe, Jonathon Shlens, and
  Zbigniew Wojna.
\newblock Rethinking the inception architecture for computer vision, 2015.

\bibitem{ham10000}
Philipp Tschandl.
\newblock {The HAM10000 dataset, a large collection of multi-source
  dermatoscopic images of common pigmented skin lesions}, 2018.

\bibitem{silver07}
Bram van Ginneken.
\newblock Sliver07, March 2019.

\bibitem{3d-ircad}
L~Soler, A~Hostettler, V~Agnus, A~Charnoz, J~Fasquel, J~Moreau, A~Osswald,
  M~Bouhadjar, and J~Marescaux.
\newblock 3d image reconstruction for comparison of algorithm database: A
  patient specific anatomical and medical image database.
\newblock {\em IRCAD, Strasbourg, France, Tech. Rep}, 1(1), 2010.

\bibitem{dropoutref}
Nitish Srivastava, Geoffrey Hinton, Alex Krizhevsky, Ilya Sutskever, and Ruslan
  Salakhutdinov.
\newblock Dropout: A simple way to prevent neural networks from overfitting.
\newblock {\em Journal of Machine Learning Research}, 15(56):1929--1958, 2014.

\bibitem{menegola2017recod}
Afonso Menegola, Julia Tavares, Michel Fornaciali, Lin~Tzy Li, Sandra Avila,
  and Eduardo Valle.
\newblock Recod titans at isic challenge 2017.
\newblock {\em arXiv preprint arXiv:1703.04819}, 2017.

\bibitem{szegedy2017inception}
Christian Szegedy, Sergey Ioffe, Vincent Vanhoucke, and Alexander~A Alemi.
\newblock Inception-v4, inception-resnet and the impact of residual connections
  on learning.
\newblock In {\em Thirty-first AAAI conference on artificial intelligence},
  2017.

\bibitem{mendoncca2013ph}
Teresa Mendon{\c{c}}a, Pedro~M Ferreira, Jorge~S Marques, Andr{\'e}~RS Marcal,
  and Jorge Rozeira.
\newblock Ph 2-a dermoscopic image database for research and benchmarking.
\newblock In {\em 2013 35th annual international conference of the IEEE
  engineering in medicine and biology society (EMBC)}, pages 5437--5440. IEEE,
  2013.

\bibitem{kawahara2018seven}
Jeremy Kawahara, Sara Daneshvar, Giuseppe Argenziano, and Ghassan Hamarneh.
\newblock Seven-point checklist and skin lesion classification using multitask
  multimodal neural nets.
\newblock {\em IEEE journal of biomedical and health informatics},
  23(2):538--546, 2018.

\bibitem{zhang2020deep}
Kai Zhang, Luc~Van Gool, and Radu Timofte.
\newblock Deep unfolding network for image super-resolution.
\newblock In {\em Proceedings of the IEEE/CVF conference on computer vision and
  pattern recognition}, pages 3217--3226, 2020.

\bibitem{sift}
David~G Lowe.
\newblock Distinctive image features from scale-invariant keypoints.
\newblock {\em International journal of computer vision}, 60(2):91--110, 2004.

\bibitem{surf}
Herbert Bay, Tinne Tuytelaars, and Luc~Van Gool.
\newblock Surf: Speeded up robust features.
\newblock In {\em European conference on computer vision}, pages 404--417.
  Springer, 2006.

\bibitem{fast}
Edward Rosten and Tom Drummond.
\newblock Machine learning for high-speed corner detection.
\newblock In {\em European conference on computer vision}, pages 430--443.
  Springer, 2006.

\bibitem{orb}
Ethan Rublee, Vincent Rabaud, Kurt Konolige, and Gary Bradski.
\newblock Orb: An efficient alternative to sift or surf.
\newblock In {\em 2011 International conference on computer vision}, pages
  2564--2571. Ieee, 2011.

\bibitem{van2008visualizing}
Laurens Van~der Maaten and Geoffrey Hinton.
\newblock Visualizing data using t-sne.
\newblock {\em Journal of machine learning research}, 9(11), 2008.

\bibitem{ckplus}
Patrick Lucey, Jeffrey~F. Cohn, Takeo Kanade, Jason Saragih, Zara Ambadar, and
  Iain Matthews.
\newblock The extended cohn-kanade dataset (ck+): A complete dataset for action
  unit and emotion-specified expression.
\newblock In {\em 2010 IEEE Computer Society Conference on Computer Vision and
  Pattern Recognition - Workshops}, pages 94--101, 2010.

\bibitem{openface}
Tadas Baltrusaitis, Amir Zadeh, Yao~Chong Lim, and Louis-Philippe Morency.
\newblock Openface 2.0: Facial behavior analysis toolkit.
\newblock In {\em 2018 13th IEEE international conference on automatic face \&
  gesture recognition (FG 2018)}, pages 59--66. IEEE, 2018.

\bibitem{riaz2020exnet}
Muhammad~Naveed Riaz, Yao Shen, Muhammad Sohail, and Minyi Guo.
\newblock Exnet: An efficient approach for emotion recognition in the wild.
\newblock {\em Sensors}, 20(4):1087, 2020.

\bibitem{ekman1978facial}
Paul Ekman and Wallace~V Friesen.
\newblock Facial action coding system.
\newblock {\em Environmental Psychology \& Nonverbal Behavior}, 1978.

\bibitem{ucar2020covidiagnosis}
Ferhat Ucar and Deniz Korkmaz.
\newblock Covidiagnosis-net: Deep bayes-squeezenet based diagnosis of the
  coronavirus disease 2019 (covid-19) from x-ray images.
\newblock {\em Medical hypotheses}, 140:109761, 2020.

\bibitem{wang2020covid}
Linda Wang, Zhong~Qiu Lin, and Alexander Wong.
\newblock Covid-net: A tailored deep convolutional neural network design for
  detection of covid-19 cases from chest x-ray images.
\newblock {\em Scientific Reports}, 10(1):1--12, 2020.

\bibitem{tangermann2012review}
Michael Tangermann, Klaus-Robert M{\"u}ller, Ad~Aertsen, Niels Birbaumer,
  Christoph Braun, Clemens Brunner, Robert Leeb, Carsten Mehring, Kai~J Miller,
  Gernot Mueller-Putz, et~al.
\newblock Review of the bci competition iv.
\newblock {\em Frontiers in neuroscience}, page~55, 2012.

\bibitem{lee2019eeg}
Min-Ho Lee, O-Yeon Kwon, Yong-Jeong Kim, Hong-Kyung Kim, Young-Eun Lee, John
  Williamson, Siamac Fazli, and Seong-Whan Lee.
\newblock Eeg dataset and openbmi toolbox for three bci paradigms: An
  investigation into bci illiteracy.
\newblock {\em GigaScience}, 8(5):giz002, 2019.

\bibitem{marek2011parkinson}
Kenneth Marek, Danna Jennings, Shirley Lasch, Andrew Siderowf, Caroline Tanner,
  Tanya Simuni, Chris Coffey, Karl Kieburtz, Emily Flagg, Sohini Chowdhury,
  et~al.
\newblock The parkinson progression marker initiative (ppmi).
\newblock {\em Progress in neurobiology}, 95(4):629--635, 2011.

\bibitem{van2016conditional}
Aaron Van~den Oord, Nal Kalchbrenner, Lasse Espeholt, Oriol Vinyals, Alex
  Graves, et~al.
\newblock Conditional image generation with pixelcnn decoders.
\newblock {\em Advances in neural information processing systems}, 29, 2016.

\bibitem{kylberg2011kylberg}
Gustaf Kylberg.
\newblock {\em Kylberg texture dataset v. 1.0}.
\newblock Centre for Image Analysis, Swedish University of Agricultural
  Sciences and~…, 2011.

\bibitem{uehara2019prototype}
Kazuki Uehara, Masahiro Murakawa, Hirokazu Nosato, and Hidenori Sakanashi.
\newblock Prototype-based interpretation of pathological image analysis by
  convolutional neural networks.
\newblock In {\em Asian Conference on Pattern Recognition}, pages 640--652.
  Springer, 2019.

\bibitem{li2018deep}
Oscar Li, Hao Liu, Chaofan Chen, and Cynthia Rudin.
\newblock Deep learning for case-based reasoning through prototypes: A neural
  network that explains its predictions.
\newblock In {\em Proceedings of the AAAI Conference on Artificial
  Intelligence}, volume~32, 2018.

\bibitem{liu2017detecting}
Yun Liu, Krishna Gadepalli, Mohammad Norouzi, George~E Dahl, Timo Kohlberger,
  Aleksey Boyko, Subhashini Venugopalan, Aleksei Timofeev, Philip~Q Nelson,
  Greg~S Corrado, et~al.
\newblock Detecting cancer metastases on gigapixel pathology images.
\newblock {\em arXiv preprint arXiv:1703.02442}, 2017.

\bibitem{ludovic2013mitosis}
Roux Ludovic, Racoceanu Daniel, Lom{\'e}nie Nicolas, Kulikova Maria, Irshad
  Humayun, Klossa Jacques, Capron Fr{\'e}d{\'e}rique, Genestie Catherine,
  et~al.
\newblock Mitosis detection in breast cancer histological images an icpr 2012
  contest.
\newblock {\em Journal of pathology informatics}, 4(1):8, 2013.

\bibitem{veta2016tumor}
Mitko Veta, Josien~PW Pluim, Nikolaos Stathonikos, Paul~J van Diest, Francisco
  Beca, and Andrew Beck.
\newblock Tumor proliferation assessment challenge 2016, miccai grand
  challenge.
\newblock 2016.

\bibitem{mnih2014recurrent}
Volodymyr Mnih, Nicolas Heess, Alex Graves, et~al.
\newblock Recurrent models of visual attention.
\newblock {\em Advances in neural information processing systems}, 27, 2014.

\bibitem{simonyan2014very}
Karen Simonyan and Andrew Zisserman.
\newblock Very deep convolutional networks for large-scale image recognition.
\newblock {\em arXiv preprint arXiv:1409.1556}, 2014.

\bibitem{lecun1998gradient}
Yann LeCun, L{\'e}on Bottou, Yoshua Bengio, and Patrick Haffner.
\newblock Gradient-based learning applied to document recognition.
\newblock {\em Proceedings of the IEEE}, 86(11):2278--2324, 1998.

\bibitem{pintelas2020explainable}
Emmanuel Pintelas, Meletis Liaskos, Ioannis~E Livieris, Sotiris Kotsiantis, and
  Panagiotis Pintelas.
\newblock Explainable machine learning framework for image classification
  problems: case study on glioma cancer prediction.
\newblock {\em Journal of imaging}, 6(6):37, 2020.

\bibitem{chen2016xgboost}
Tianqi Chen and Carlos Guestrin.
\newblock Xgboost: A scalable tree boosting system.
\newblock In {\em Proceedings of the 22nd acm sigkdd international conference
  on knowledge discovery and data mining}, pages 785--794, 2016.

\bibitem{rajaraman2018pre}
Sivaramakrishnan Rajaraman, Sameer~K Antani, Mahdieh Poostchi, Kamolrat
  Silamut, Md~A Hossain, Richard~J Maude, Stefan Jaeger, and George~R Thoma.
\newblock Pre-trained convolutional neural networks as feature extractors
  toward improved malaria parasite detection in thin blood smear images.
\newblock {\em PeerJ}, 6:e4568, 2018.

\bibitem{huang2008wavelet}
Ke~Huang and Selin Aviyente.
\newblock Wavelet feature selection for image classification.
\newblock {\em IEEE Transactions on Image Processing}, 17(9):1709--1720, 2008.

\bibitem{mirjalili2016whale}
Seyedali Mirjalili and Andrew Lewis.
\newblock The whale optimization algorithm.
\newblock {\em Advances in engineering software}, 95:51--67, 2016.

\bibitem{9093853}
Muhammad Umer, Saima Sadiq, Muhammad Ahmad, Saleem Ullah, Gyu~Sang Choi, and
  Arif Mehmood.
\newblock A novel stacked cnn for malarial parasite detection in thin blood
  smear images.
\newblock {\em IEEE Access}, 8:93782--93792, 2020.

\bibitem{rajaraman2019performance}
Sivaramakrishnan Rajaraman, Stefan Jaeger, and Sameer~K Antani.
\newblock Performance evaluation of deep neural ensembles toward malaria
  parasite detection in thin-blood smear images.
\newblock {\em PeerJ}, 7:e6977, 2019.

\bibitem{8697909}
A.~Sai~Bharadwaj Reddy and D.~Sujitha Juliet.
\newblock Transfer learning with resnet-50 for malaria cell-image
  classification.
\newblock pages 0945--0949, 2019.

\bibitem{bray2018global}
Freddie Bray, Jacques Ferlay, Isabelle Soerjomataram, Rebecca~L Siegel,
  Lindsey~A Torre, and Ahmedin Jemal.
\newblock Global cancer statistics 2018: Globocan estimates of incidence and
  mortality worldwide for 36 cancers in 185 countries.
\newblock {\em CA: a cancer journal for clinicians}, 68(6):394--424, 2018.

\bibitem{heath00}
M~Heath, K~Bowyer, D~Kopans, R~Moore, and P~Kegelmeyer.
\newblock The digital database for screening mammography.
\newblock In {\em Proceedings of the Fifth International Workshop on Digital
  Mammography}, pages 212--218, 2020.

\bibitem{lee2017curated}
Rebecca~Sawyer Lee, Francisco Gimenez, Assaf Hoogi, Kanae~Kawai Miyake, Mia
  Gorovoy, and Daniel~L Rubin.
\newblock A curated mammography data set for use in computer-aided detection
  and diagnosis research.
\newblock {\em Scientific data}, 4(1):1--9, 2017.

\bibitem{8363547}
Noel C.~F. Codella, David Gutman, M.~Emre Celebi, Brian Helba, Michael~A.
  Marchetti, Stephen~W. Dusza, Aadi Kalloo, Konstantinos Liopyris, Nabin
  Mishra, Harald Kittler, and Allan Halpern.
\newblock Skin lesion analysis toward melanoma detection: A challenge at the
  2017 international symposium on biomedical imaging (isbi), hosted by the
  international skin imaging collaboration (isic).
\newblock In {\em 2018 IEEE 15th International Symposium on Biomedical Imaging
  (ISBI 2018)}, pages 168--172, 2018.

\bibitem{litjens2017survey}
Geert Litjens, Thijs Kooi, Babak~Ehteshami Bejnordi, Arnaud Arindra~Adiyoso
  Setio, Francesco Ciompi, Mohsen Ghafoorian, Jeroen~Awm Van Der~Laak, Bram
  Van~Ginneken, and Clara~I S{\'a}nchez.
\newblock A survey on deep learning in medical image analysis.
\newblock {\em Medical image analysis}, 42:60--88, 2017.

\bibitem{avanzo2020machine}
Michele Avanzo, Lise Wei, Joseph Stancanello, Martin Vallieres, Arvind Rao,
  Olivier Morin, Sarah~A Mattonen, and Issam El~Naqa.
\newblock Machine and deep learning methods for radiomics.
\newblock {\em Medical physics}, 47(5):e185--e202, 2020.

\bibitem{hatt2019machine}
Mathieu Hatt, Chintan Parmar, Jinyi Qi, and Issam El~Naqa.
\newblock Machine (deep) learning methods for image processing and radiomics.
\newblock {\em IEEE Transactions on Radiation and Plasma Medical Sciences},
  3(2):104--108, 2019.

\bibitem{ribeiro2016should}
Marco~Tulio Ribeiro, Sameer Singh, and Carlos Guestrin.
\newblock " why should i trust you?" explaining the predictions of any
  classifier.
\newblock In {\em Proceedings of the 22nd ACM SIGKDD international conference
  on knowledge discovery and data mining}, pages 1135--1144, 2016.

\bibitem{li2021normalization}
Yajun Li, Guoqiang Han, Xiaomei Wu, Zhen~Hui Li, Ke~Zhao, Zhiping Zhang, Zaiyi
  Liu, and Changhong Liang.
\newblock Normalization of multicenter ct radiomics by a generative adversarial
  network method.
\newblock {\em Physics in Medicine \& Biology}, 66(5):055030, 2021.

\bibitem{johnson2007adjusting}
W~Evan Johnson, Cheng Li, and Ariel Rabinovic.
\newblock Adjusting batch effects in microarray expression data using empirical
  bayes methods.
\newblock {\em Biostatistics}, 8(1):118--127, 2007.

\bibitem{pmlr-v37-ioffe15}
Sergey Ioffe and Christian Szegedy.
\newblock Batch normalization: Accelerating deep network training by reducing
  internal covariate shift.
\newblock In Francis Bach and David Blei, editors, {\em Proceedings of the 32nd
  International Conference on Machine Learning}, volume~37 of {\em Proceedings
  of Machine Learning Research}, pages 448--456, Lille, France, 07--09 Jul
  2015. PMLR.

\bibitem{ilse2018attention}
Maximilian Ilse, Jakub Tomczak, and Max Welling.
\newblock Attention-based deep multiple instance learning.
\newblock In {\em International conference on machine learning}, pages
  2127--2136. PMLR, 2018.

\bibitem{combalia2019bcn20000}
Marc Combalia, Noel~CF Codella, Veronica Rotemberg, Brian Helba, Veronica
  Vilaplana, Ofer Reiter, Cristina Carrera, Alicia Barreiro, Allan~C Halpern,
  Susana Puig, et~al.
\newblock Bcn20000: Dermoscopic lesions in the wild.
\newblock {\em arXiv preprint arXiv:1908.02288}, 2019.

\bibitem{tschandl2018ham10000}
Philipp Tschandl, Cliff Rosendahl, and Harald Kittler.
\newblock The ham10000 dataset, a large collection of multi-source
  dermatoscopic images of common pigmented skin lesions.
\newblock {\em Scientific data}, 5(1):1--9, 2018.

\bibitem{6610779}
Teresa Mendonça, Pedro~M. Ferreira, Jorge~S. Marques, André R.~S. Marcal, and
  Jorge Rozeira.
\newblock Ph2 - a dermoscopic image database for research and benchmarking.
\newblock In {\em 2013 35th Annual International Conference of the IEEE
  Engineering in Medicine and Biology Society (EMBC)}, pages 5437--5440, 2013.

\bibitem{8333693}
Jeremy Kawahara, Sara Daneshvar, Giuseppe Argenziano, and Ghassan Hamarneh.
\newblock Seven-point checklist and skin lesion classification using multitask
  multimodal neural nets.
\newblock {\em IEEE Journal of Biomedical and Health Informatics},
  23(2):538--546, 2019.

\bibitem{ogurtsova2017idf}
Katherine Ogurtsova, JD~da~Rocha~Fernandes, Y~Huang, Ute Linnenkamp,
  L~Guariguata, Nam~H Cho, David Cavan, JE~Shaw, and LE~Makaroff.
\newblock Idf diabetes atlas: Global estimates for the prevalence of diabetes
  for 2015 and 2040.
\newblock {\em Diabetes research and clinical practice}, 128:40--50, 2017.

\bibitem{yu2021reconfigisp}
Ke~Yu, Zexian Li, Yue Peng, Chen~Change Loy, and Jinwei Gu.
\newblock Reconfigisp: Reconfigurable camera image processing pipeline, 2021.

\bibitem{ting2019deep}
Daniel~SW Ting, Lily Peng, Avinash~V Varadarajan, Pearse~A Keane, Philippe~M
  Burlina, Michael~F Chiang, Leopold Schmetterer, Louis~R Pasquale, Neil~M
  Bressler, Dale~R Webster, et~al.
\newblock Deep learning in ophthalmology: the technical and clinical
  considerations.
\newblock {\em Progress in retinal and eye research}, 72:100759, 2019.

\bibitem{gilpin2018explaining}
Leilani~H Gilpin, David Bau, Ben~Z Yuan, Ayesha Bajwa, Michael Specter, and
  Lalana Kagal.
\newblock Explaining explanations: An overview of interpretability of machine
  learning.
\newblock In {\em 2018 IEEE 5th International Conference on data science and
  advanced analytics (DSAA)}, pages 80--89. IEEE, 2018.

\bibitem{russell2015research}
Stuart Russell, Daniel Dewey, and Max Tegmark.
\newblock Research priorities for robust and beneficial artificial
  intelligence.
\newblock {\em Ai Magazine}, 36(4):105--114, 2015.

\bibitem{goodman2017european}
Bryce Goodman and Seth Flaxman.
\newblock European union regulations on algorithmic decision-making and a
  “right to explanation”.
\newblock {\em AI magazine}, 38(3):50--57, 2017.

\bibitem{mixmatch}
David Berthelot, Nicholas Carlini, Ian Goodfellow, Nicolas Papernot, Avital
  Oliver, and Colin~A Raffel.
\newblock Mixmatch: A holistic approach to semi-supervised learning.
\newblock {\em Advances in neural information processing systems}, 32, 2019.

\bibitem{gal2016dropout}
Yarin Gal and Zoubin Ghahramani.
\newblock Dropout as a bayesian approximation: Representing model uncertainty
  in deep learning.
\newblock In {\em international conference on machine learning}, pages
  1050--1059. PMLR, 2016.

\bibitem{van2020simple}
Joost van Amersfoort, Lewis Smith, Yee~Whye Teh, and Yarin Gal.
\newblock Simple and scalable epistemic uncertainty estimation using a single
  deep deterministic neural network.
\newblock 2020.

\end{thebibliography}

\end{document}